\documentclass{article} 
\usepackage[final]{colm2026_conference}

\usepackage{microtype}
\usepackage{hyperref}
\usepackage{url}
\usepackage{booktabs}
\usepackage[T1]{fontenc}
\usepackage{adjustbox}
\usepackage{kotex}
\usepackage{graphicx}
\usepackage{amssymb}
\usepackage{booktabs}
\usepackage{multirow}
\usepackage{xcolor}
\usepackage{color,soul}
\usepackage{amsmath}
\usepackage{listings, listings-rust}
\usepackage[table]{xcolor}
\usepackage{pifont}
\usepackage{enumitem}
\usepackage{tabularx}   
\usepackage{makecell}   
\usepackage{caption}    
\usepackage{tcolorbox}
\usepackage{fvextra} 

\usepackage[pass]{geometry}

\usepackage{lineno}

\definecolor{darkblue}{rgb}{0, 0, 0.5}
\hypersetup{colorlinks=true, citecolor=darkblue, linkcolor=darkblue, urlcolor=darkblue}
\definecolor{successcolor}{RGB}{235, 250, 235}
\definecolor{failcolor}{RGB}{250, 235, 235}

\newcommand{\cmark}{\textcolor{green!50!black}{\checkmark}}
\newcommand{\xmark}{\textcolor{red!50!black}{\ding{55}}} 

\newcommand{\blueh}{\textcolor{blue}}
\newcommand{\redh}{\textcolor{red}}

\title{KoSimpleQA: A Korean Factuality Benchmark\\with an Analysis of Reasoning LLMs}

\author{Donghyeon Ko\textsuperscript{$\ast$1},
Kyubyung Chae\textsuperscript{$\ast$2},
Yeguk Jin\textsuperscript{1},
Byungwook Lee\textsuperscript{1},
Chansong Jo\textsuperscript{1}, \\
\bf Sookyo In\textsuperscript{1},
Jaehong Lee\textsuperscript{1},
Taesup Kim\textsuperscript{$\dagger$2},
Donghyun Kwak\textsuperscript{$\dagger$1}
\\
\\
\textsuperscript{1}Naver Cloud,
\textsuperscript{2}Graduate School of Data Science, Seoul National University
\\
\small{
\{donghyeon.ko, donghyun.kwak\}@navercorp.com,
\{kyubyung.chae, taesup.kim\}@snu.ac.kr
}
}
\begingroup

\footnotetext{$^\ast$ Equal contribution. $^\dagger$ Co-corresponding authors.}
\endgroup

\begin{document}

\ifcolmsubmission
\linenumbers
\fi

\maketitle

\begin{abstract}
We present \textbf{Korean SimpleQA (KoSimpleQA)}, a benchmark for evaluating factuality in large language models (LLMs) with a focus on Korean cultural knowledge. KoSimpleQA is designed to be challenging yet easy to grade, consisting of 938 short, fact-seeking questions with unambiguous answers. We conduct a comprehensive evaluation across a diverse set of open-source LLMs of varying sizes that support Korean, and find that even the strongest model generates correct answer only 31.6\% of the time, underscoring the challenging nature of KoSimpleQA. Notably, performance rankings on KoSimpleQA differ substantially from those on the English SimpleQA, highlighting the unique value of our dataset. Furthermore, we observe that reasoning helps mitigate the cross-lingual knowledge gap in LLMs, which refers to disparities in their ability to manifest knowledge across languages.
KoSimpleQA can be found at \url{https://github.com/naver-ai/KoSimpleQA}.
\end{abstract}

\section{Introduction}
\vspace{-0.2cm}
Large language models (LLMs) are increasingly deployed in applications that demand high factual reliability, ranging from information access to education and cultural content delivery. However, despite recent progress, LLMs still frequently generate factually incorrect outputs, a phenomenon widely known as \textit{hallucination} \citep{banerjee2025llms}. This has motivated the development of benchmarks that directly measure factual reliability, such as SimpleQA \citep{wei2024measuring} and Chinese SimpleQA \citep{he2024chinese}, which evaluate models on short, fact-seeking questions with unambiguous answers.

However, existing benchmarks primarily focus on English and Chinese, while no equivalent benchmark exists for Korean. Simply translating existing benchmarks such as SimpleQA into Korean is not sufficient, since many of their questions are rooted in Anglophone cultural contexts and do not meaningfully assess models trained primarily on Korean data. Evaluating LLMs in a particular language requires not only linguistic competence but also an understanding of the cultural knowledge associated with that language community. 

\begin{figure}
    \centering
    \includegraphics[scale=0.6]{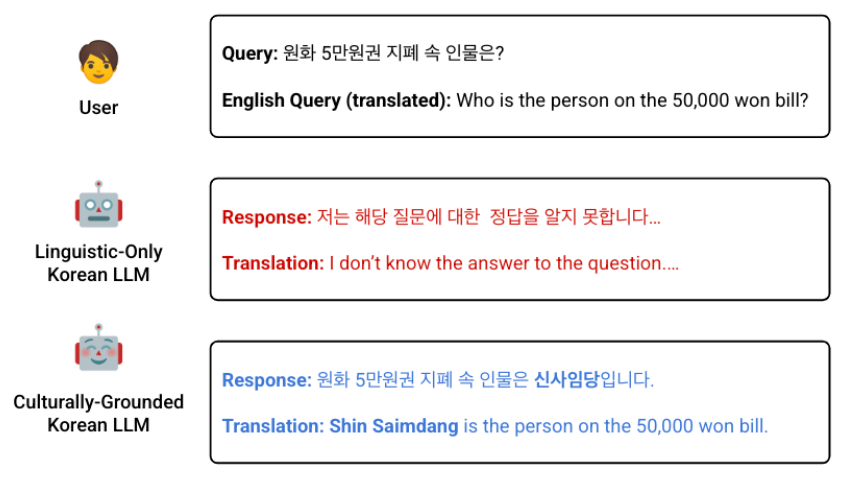}
    \caption{The importance of cultural knowledge for LLMs within a specific language community: an LLM with only linguistic competence responds fluently in Korean but fails, whereas one with Korean cultural knowledge correctly answers the question “Who is the person on the 50,000 won bill?”}
    \vspace{-0.2cm}
    \label{fig:motivation}
\end{figure}

Figure \ref{fig:motivation} illustrates this distinction: a model with only linguistic competence can respond fluently in Korean but still fails on a culturally grounded question, whereas a model equipped with Korean cultural knowledge produces the correct answer. The release of Chinese SimpleQA further demonstrates that benchmarks need to be adapted to each language community’s cultural context, rather than assuming a single benchmark can generalize across languages. In the same spirit, we propose \textbf{Korean SimpleQA (KoSimpleQA)}, which fills this gap for Korean by providing a benchmark centered on Korean cultural knowledge. Examples of KoSimpleQA are provided in Appendix~\ref{app:stat}.

Following the design principles of SimpleQA-style benchmarks, KoSimpleQA is challenging yet easy to grade, consisting of short, fact-seeking questions with unambiguous answers. We performed multiple rounds of human validation to 
control the quality of the dataset. We then systematically evaluated a broad range of open-source LLMs that support Korean, spanning different model families and parameter scales. Despite this diversity, no model generates correct answers more than  31.6\% of the time, underscoring the difficulty of the benchmark and suggesting substantial room for further advancement in non-English, low-resource language settings. Moreover, performance rankings on KoSimpleQA diverge substantially from those on English SimpleQA, highlighting the distinct cultural dimension it captures and emphasizing the importance of localized, culturally grounded evaluations.

In addition, we analyze the behavior of recent reasoning LLMs on KoSimpleQA. While reasoning has been emphasized as a core capability of frontier models, its role in factual QA has not been systematically examined. Specifically, we find that leveraging a model’s reasoning ability can 
mitigate cross-lingual knowledge gap observed in non-reasoning responses, where the model fails to express knowledge it already possesses when the query language changes, and can help elicit its latent knowledge. In such cases, reasoning serves as a mechanism for bridging the gap between internal knowledge and its correct verbalization in factual question answering.

Our contributions are threefold:
\vspace{-0.2cm}
\begin{itemize}
    \item We release \textbf{Korean SimpleQA (KoSimpleQA)}, the first benchmark targeting factuality in Korean cultural contexts.
    \item We provide a comprehensive evaluation across diverse LLMs, establishing new baselines and showing differences from SimpleQA.
    \item We additionally analyze reasoning models, offering preliminary insights into the role of reasoning in factual QA.
\end{itemize}

\vspace{-0.2cm}
\section{Related Work}
\vspace{-0.2cm}
\subsection{Korean Factuality Benchmarks}\label{subsec:korean_benchmarks}

Several benchmarks have been proposed to evaluate factuality while incorporating Korean cultural knowledge, including KorNAT~\citep{lee-etal-2024-kornat}, CLIcK~\citep{kim2024click}, HAE-RAE~\citep{son2024haerae}, and KULTURE~\citep{wang2024kulture}. However, these benchmarks share many of the limitations observed in traditional factuality benchmarks such as TruthfulQA~\citep{lin2022truthfulqa} and CommonsenseQA~\citep{talmor-etal-2019-commonsenseqa}.

First, their relatively low difficulty has led to performance saturation, making it difficult to discriminate between top-tier models. This issue may stem from insufficient consideration of difficulty during data collection, as well as inherent limitations of the multiple-choice format. In multiple-choice settings, the answer candidate space is constrained, allowing models to guess correctly or rely on elimination strategies, which undermines the reliability of the evaluation~\citep{tam-etal-2025-none, singh2025optionspitfallsmultiplechoicequestions, li-etal-2024-multiple, balepur-etal-2025-best}.

Second, these benchmarks do not adequately account for the desirable behavior of models abstaining when uncertain~\citep{madhusudhan2024llmsknowanswerinvestigating}. From a hallucination perspective, a model admitting uncertainty (“I don’t know”) is preferable to providing an incorrect answer. 

However, existing factuality benchmarks often treat both abstentions and incorrect answers as equally wrong. Furthermore, the multiple-choice format inherently excludes abstention by forcing the model to select one of the provided options, thereby failing to capture this important aspect of model behavior.

\vspace{-0.2cm}
\subsection{SimpleQA and Cultural Adaptation}\label{subsec:factuality_adaptation}

SimpleQA~\citep{wei2024measuring} addresses the limitations of traditional factuality benchmarks by presenting a significant challenge even to top-tier LLMs like GPT-4o~\citep{hurst2024gpt}. This difficulty is ensured through a rigorous adversarial filtering process: human annotators were instructed to regenerate questions if a suite of four OpenAI models consistently answered them correctly. This ensures that the dataset consists only of non-trivial questions. To overcome the limitations of multiple-choice formats, SimpleQA adopts an open-ended (free-form) question format while carefully designing questions to elicit concise and uniquely identifiable answers, while still enabling reliable evaluation. Furthermore, SimpleQA introduces a new evaluation framework that prioritizes reliability, specifically measuring a model's ability to abstain from answering (``know what it knows") rather than hallucinating when uncertain.

Despite its effectiveness, the original SimpleQA primarily targets English-centric knowledge, leaving a gap in our understanding of LLM capabilities within diverse cultural and linguistic landscapes. Recent efforts have sought to extend this paradigm into multilingual or localized contexts. For instance, KoLasSimpleQA~\citep{jiang2025evaluating} extends SimpleQA to nine languages, marking a valuable step toward multilingual factuality evaluation. However, it faces several limitations: the dataset remains publicly unavailable, constraining reproducibility; the sample size per language is relatively small; and its difficulty filtering is less rigorous than that of the original SimpleQA. Conversely, Chinese SimpleQA~\citep{he2024chinese} focuses rigorously on Chinese-specific knowledge, revealing substantial model ranking discrepancies compared to the original benchmark, with localized community models often outperforming frontier global models.

Motivated by this shift toward localized factuality and the need for a rigorous, open-source benchmark for Korean, we introduce KoSimpleQA, the first factuality benchmark adopting the SimpleQA paradigm for Korean cultural knowledge. Notably, while Chinese SimpleQA was constructed using machine-generated data, KoSimpleQA aligns with the original SimpleQA in prioritizing human-curated QA pairs to ensure high factual reliability and to capture cultural context.

\vspace{-0.2cm}
\section{KoSimpleQA}
\vspace{-0.2cm}
KoSimpleQA is constructed through human annotation guided by carefully designed instructions (Section \ref{subsec:data_collection}), followed by multiple rounds of verification (Section \ref{subsec:quality_control}). This methodology ensures a non-trivial benchmark consisting of questions with a single, well-defined answer, enabling reliable evaluation of the factual accuracy of LLMs within the Korean cultural context. The evaluation metrics used in this benchmark are described in Section~\ref{subsec:eval_metrics}.

\vspace{-0.2cm}
\subsection{Data Collection}\label{subsec:data_collection}
We construct a dataset comprising 938 original, human-authored knowledge-seeking questions.
We recruit annotators through two Korean crowdsourcing platforms, Selectstar\footnote{\url{https://selectstar.ai/}} and Flitto\footnote{\url{https://www.flitto.com/}}. 
All human annotators are required to pass assessments designed by our data partners to verify their capability in producing high-quality data. Qualified annotators are instructed to craft fact-based questions grounded in Korean cultural contexts. A detailed description of the annotator instructions and additional annotation statistics are provided in Appendix~\ref{app:guideline}.

\vspace{-0.2cm}
\paragraph{Cultural Grounding.}


While many cross-lingual evaluations rely on translated versions of English-centric datasets~\citep{artetxe-etal-2020-cross, bandarkar-etal-2024-belebele}, a core objective of KoSimpleQA is to capture the nuance of Korean context. We define ``culture'' broadly, extending beyond traditional heritage (e.g., history, literature) to encompass modern socio-cultural trends and contemporary interests unique to native speakers. This scope ensures that the benchmark covers diverse topics of local significance, ranging from traditional holidays to details of popular modern variety shows (e.g., \textit{I Live Alone}). Consequently, while keeping the question format simple, the content necessitates specific parametric knowledge of the Korean cultural sphere.

\vspace{-0.2cm}
\paragraph{Difficulty requirement.} Each question have to be sufficiently challenging such that at least one of the four strong closed-source LLMs we selected (HCX-005 \citep{yoo2024hyperclova}, gpt-4o-2024-05-13 \citep{hurst2024gpt}, gemini-2.0-flash\footnote{\url{https://blog.google/technology/google-deepmind/google-gemini-ai-update-december-2024/}}, and claude-3-5-sonnet-20240620\footnote{\url{https://www.anthropic.com/news/claude-3-5-sonnet}}) failed to produce the correct answer. This criterion guaranteed that KoSimpleQA poses meaningful challenges even for current high-performing systems.

\vspace{-0.2cm}
\paragraph{Unambiguous answers.} Every question is required to admit a unique and definitive short answer. For example, a question regarding the symbolic meanings of the white background in the South Korean flag has three possible answers (i.e., brightness, purity, and national spirit). In this case, we specified that \textit{all three} should be provided, thereby removing ambiguity. This unambiguity facilitates the verification of the answer.

\vspace{-0.2cm}
\paragraph{Temporal constraint.} 
To mitigate evaluation noise caused by the dynamic nature of real-world information, annotators are instructed to create items that could be answered with knowledge available up to December 31, 2023. This constraint is particularly vital for recurring events where the \textit{latest} status changes periodically. For instance, a question inquiring about the ``real name of the Final MVP of the League of Legends World Championship 2023'' is permissible and unambiguous because the tournament is concluded and the outcome is fully settled before our cutoff date. By establishing this explicit deadline, we ensure that every item remains a stable benchmark, insulated from the influence of subsequent real-world developments.

\begin{figure}[t!]
    \centering
    \includegraphics[width=0.8\columnwidth]{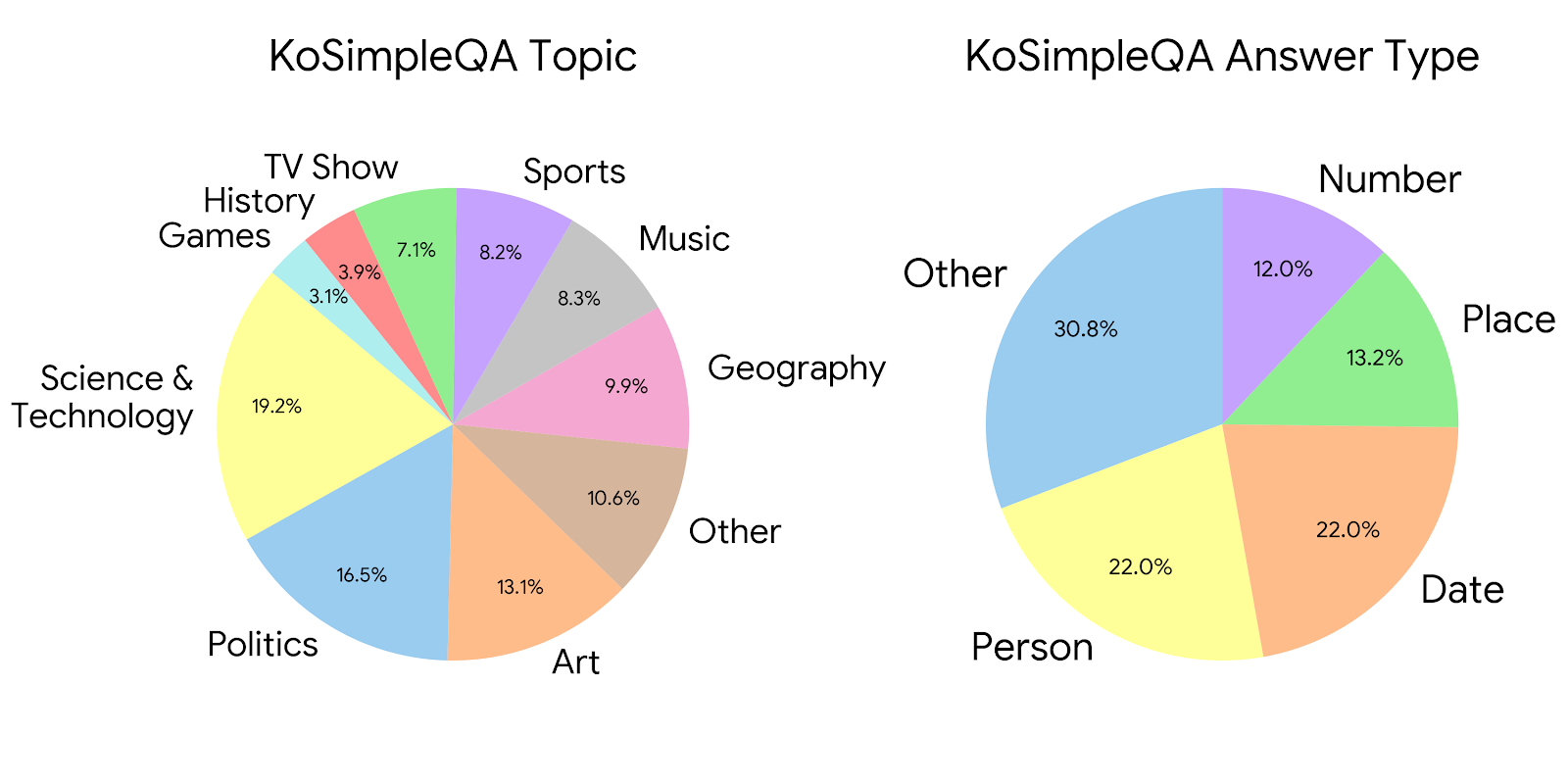}
    \vspace{-0.4cm}
    \caption{Distribution of topics and answer types in KoSimpleQA. 
    The left chart shows the breakdown by topic category, and the right chart shows the distribution by answer type.} 
    \label{fig:categories}
    \vspace{-0.2cm} 
\end{figure}

\vspace{-0.2cm}
\paragraph{Topics and Diversity.}
We adopt the topic categories defined in SimpleQA. We instruct human annotators to create data in a way that would elicit a diverse range of answer types.
Figure \ref{fig:categories} illustrates the resulting distribution of topic categories and answer types. The dataset covers a comprehensive range of subjects without skewing towards a single domain.

\vspace{-0.2cm}
\paragraph{Source verification.} 
Annotators are instructed to record up to four reliable source links supporting each answer at the time of question creation. These references are subsequently utilized during the quality control phase to verify factual accuracy and ensure reproducibility. By explicitly linking each answer to verifiable evidence, we minimize the likelihood of annotation errors and enhance transparency in dataset construction.

\vspace{-0.2cm}
\subsection{Quality Control}\label{subsec:quality_control}
\vspace{-0.2cm}
To guarantee the factual reliability and clarity of KoSimpleQA, we implement a two-stage verification process incorporating an iterative feedback loop. 
First, we employ a cross-platform validation strategy: instances generated by one platform are cross-validated by annotators from the other to identify initial discrepancies. Second, a manual expert review is conducted by NLP researchers to enforce the rigorous standards defined in Section \ref{subsec:data_collection}.

During this stage, we address various source of errors, including typographical errors, ambiguity, and time-sensitiveness. In particular, many cases related to clarity and time sensitivity are non-trivial and often overlooked without domain expertise in NLP, leading to substantial refinements at this stage. For example, some queries asking for dates are revised to explicitly request request an appropriate level of temporal granularity (e.g., year, month, and day as needed), and questions whose answers can vary over time are reformulated to specify a particular temporal reference point.

\vspace{-0.2cm}
\subsection{Evaluation Metrics}\label{subsec:eval_metrics}

\vspace{-0.2cm}
\paragraph{Standard Metrics.}
We adopt the standard metric definitions from SimpleQA~\citep{wei2024measuring}, which utilize evaluation results from an LLM judge. Following the initial release of the official implementation (\texttt{simple-evals}\footnote{\url{https://github.com/openai/simple-evals}}), we employ GPT-4o as our judge model. However, we refine the \texttt{NOT\_ATTEMPTED} criteria to prevent models from receiving credit for invalid refusals. These refinements are reflected in our evaluation prompts to ensure alignment between the metric logic and the automated grader. The metrics are defined as follows:
\begin{itemize}[nosep, leftmargin=1em]
    \item \textbf{Correct (CO)}: The proportion of predictions that fully cover the reference answer without contradicting it.
    \item \textbf{Not Attempted (NA)}: The proportion of responses that explicitly refuse to answer or state ignorance (e.g., ``I don't know''). Crucially, unlike the original relaxed evaluation, we do not classify vague responses or false refusals (e.g., claiming that no record exists, even though such information is actually documented somewhere on the web) as NA; these are penalized as Incorrect. Detailed discussion on this classification and the corresponding evaluation prompts are provided in Appendix~\ref{app:metrics_and_eval_prompts}.
    \item \textbf{Incorrect (IN)}: The proportion of predictions that are not correct.    
    \item \textbf{Correct Given Attempted (CGA)}: The proportion of accurately answered questions among those that were attempted.
    \item \textbf{F-score}: The harmonic mean of CO and CGA, providing a balanced measure of precision and completeness.
\end{itemize}

\vspace{-0.2cm}
\paragraph{Calibration Metrics.}

Beyond assessing the model’s propensity to refuse unknown questions (NA), a robust evaluation must account for its metacognitive reliability when it chooses to respond. This is critical because a model that provides \textit{confidently incorrect} answers is fundamentally less reliable than one that acknowledges its uncertainty. We thus evaluate calibration, which we measure by eliciting the model's verbalized confidence. Following \citet{wei2024measuring} and \citet{he2024chinese}, we instruct models to provide a confidence level from 0 to 100 alongside each response (see Appendix~\ref{app:cal_prompt} for the prompt). 

Prior work~\citep{wei2024measuring, he2024chinese} primarily analyzed calibration plots, without detailed quantitative evaluations. 
In contrast, we adopt \textbf{Expected Calibration Error (ECE)} \citep{ece2015} that is often employed \citep{xiong2024can, geng-etal-2024-survey, tian-etal-2023-just, yoon2025reasoning} to enable quantitative analysis of calibration. ECE quantifies the average discrepancy between accuracy and predicted confidence across binned intervals, weighted by sample size. Importantly, these metrics are calculated exclusively on attempted questions to decouple the reliability of the model’s uncertainty estimation from its refusal behavior.

\begin{table*}[t!]
\centering
\resizebox{0.9\textwidth}{!}{%
\begin{tabular}{l|ccccc|c}
\toprule
\textbf{Models} & \textbf{CO} $\uparrow$ & \textbf{NA} & \textbf{IN} $\downarrow$ & \textbf{CGA} $\uparrow$ & \textbf{F-score} $\uparrow$ & \textbf{ECE} $\downarrow$ \\
\midrule\midrule
\multicolumn{7}{c}{\textit{Korean Community LLMs}} \\
\midrule
EXAONE 4.0-1.2B & 1.6 {\small $\pm$ 0.6} & 4.7 {\small $\pm$ 1.2} & 93.7 {\small $\pm$ 1.6} & 1.7 {\small $\pm$ 0.6} & 1.6 {\small $\pm$ 0.6} & 80.5 {\small $\pm$ 0.8} \\
EXAONE 4.0-32B & 17.2 {\small $\pm$ 0.4} & 1.2 {\small $\pm$ 0.1} & 81.6 {\small $\pm$ 0.5} & 17.4 {\small $\pm$ 0.4} & 17.3 {\small $\pm$ 0.4} & 68.4 {\small $\pm$ 0.3} \\
\midrule
HCX SEED-0.5B & 3.6 {\small $\pm$ 1.1} & 2.3 {\small $\pm$ 0.7} & 94.1 {\small $\pm$ 1.8} & 3.6 {\small $\pm$ 1.1} & 3.6 {\small $\pm$ 1.1} & 75.3 {\small $\pm$ 2.5} \\
HCX SEED-1.5B & 3.3 {\small $\pm$ 0.4} & 2.6 {\small $\pm$ 0.6} & 94.2 {\small $\pm$ 0.9} & 3.4 {\small $\pm$ 0.4} & 3.3 {\small $\pm$ 0.4} & 90.9 {\small $\pm$ 0.9} \\
HCX SEED-3B & 8.7 {\small $\pm$ 0.6} & 5.4 {\small $\pm$ 1.0} & 85.8 {\small $\pm$ 1.5} & 9.2 {\small $\pm$ 0.7} & 9.0 {\small $\pm$ 0.6} & 81.2 {\small $\pm$ 0.8} \\
HCX SEED-14B & \textbf{31.6} {\small $\pm$ 0.9} & 6.1 {\small $\pm$ 0.2} & \textbf{62.3} {\small $\pm$ 1.1} & \textbf{33.6} {\small $\pm$ 1.0} & \textbf{32.6} {\small $\pm$ 0.9} & \textbf{61.5} {\small $\pm$ 0.7} \\
\midrule
Kanana 1.5-2.1B & 14.6 {\small $\pm$ 0.9} & 9.3 {\small $\pm$ 0.6} & 76.1 {\small $\pm$ 0.9} & 16.1 {\small $\pm$ 1.0} & 15.3 {\small $\pm$ 0.9} & 78.8 {\small $\pm$ 0.5} \\
Kanana 1.5-8B & 27.7 {\small $\pm$ 0.5} & 1.2 {\small $\pm$ 0.3} & 71.0 {\small $\pm$ 0.3} & 28.1 {\small $\pm$ 0.4} & 27.9 {\small $\pm$ 0.5} & 64.1 {\small $\pm$ 0.6} \\
\midrule
VARCO-8B & 5.3 {\small $\pm$ 0.4} & 4.8 {\small $\pm$ 0.3} & 89.9 {\small $\pm$ 0.3} & 5.5 {\small $\pm$ 0.4} & 5.4 {\small $\pm$ 0.4} & 86.7 {\small $\pm$ 1.1} \\
\midrule
\multicolumn{7}{c}{\textit{Multilingual LLMs Supporting Korean}} \\
\midrule
Gemma 3-1B & 2.2 {\small $\pm$ 0.1} & 0.9 {\small $\pm$ 0.1} & 97.0 {\small $\pm$ 0.1} & 2.2 {\small $\pm$ 0.1} & 2.2 {\small $\pm$ 0.1} & 83.6 {\small $\pm$ 0.3} \\
Gemma 3-4B & 6.6 {\small $\pm$ 0.2} & 0.4 {\small $\pm$ 0.2} & 93.0 {\small $\pm$ 0.1} & 6.6 {\small $\pm$ 0.2} & 6.6 {\small $\pm$ 0.2} & 92.2 {\small $\pm$ 0.1} \\
Gemma 3-12B & 17.2 {\small $\pm$ 0.3} & 0.1 {\small $\pm$ 0.1} & 82.7 {\small $\pm$ 0.3} & 17.2 {\small $\pm$ 0.3} & 17.2 {\small $\pm$ 0.3} & 80.8 {\small $\pm$ 0.3} \\
Gemma 3-27B & 28.3 {\small $\pm$ 0.1} & 0.0 {\small $\pm$ 0.0} & 71.7 {\small $\pm$ 0.1} & 28.3 {\small $\pm$ 0.1} & 28.3 {\small $\pm$ 0.1} & 70.0 {\small $\pm$ 0.1} \\
\midrule
Llama 3.1-8B & 5.3 {\small $\pm$ 0.7} & 2.1 {\small $\pm$ 1.2} & 92.6 {\small $\pm$ 1.9} & 5.4 {\small $\pm$ 0.8} & 5.3 {\small $\pm$ 0.7} & 83.1 {\small $\pm$ 2.8} \\
Llama 3.1-70B & 15.6 {\small $\pm$ 0.3} & 0.2 {\small $\pm$ 0.1} & 84.1 {\small $\pm$ 0.4} & 15.7 {\small $\pm$ 0.3} & 15.7 {\small $\pm$ 0.3} & 71.4 {\small $\pm$ 4.2} \\
\midrule
Qwen 3-0.6B & 0.8 {\small $\pm$ 0.3} & 0.7 {\small $\pm$ 0.2} & 98.4 {\small $\pm$ 0.4} & 0.8 {\small $\pm$ 0.3} & 0.8 {\small $\pm$ 0.3} & 85.0 {\small $\pm$ 1.8} \\
Qwen 3-1.7B & 1.6 {\small $\pm$ 0.4} & 0.7 {\small $\pm$ 0.2} & 97.7 {\small $\pm$ 0.6} & 1.6 {\small $\pm$ 0.4} & 1.6 {\small $\pm$ 0.4} & 91.1 {\small $\pm$ 0.5} \\
Qwen 3-4B & 3.2 {\small $\pm$ 0.3} & 1.0 {\small $\pm$ 0.3} & 95.9 {\small $\pm$ 0.5} & 3.2 {\small $\pm$ 0.3} & 3.2 {\small $\pm$ 0.3} & 88.9 {\small $\pm$ 0.2} \\
Qwen 3-8B & 5.3 {\small $\pm$ 0.4} & 0.7 {\small $\pm$ 0.4} & 94.0 {\small $\pm$ 0.3} & 5.3 {\small $\pm$ 0.4} & 5.3 {\small $\pm$ 0.4} & 89.9 {\small $\pm$ 0.1} \\
Qwen 3-14B & 9.0 {\small $\pm$ 0.9} & 1.1 {\small $\pm$ 0.3} & 89.9 {\small $\pm$ 0.9} & 9.1 {\small $\pm$ 0.9} & 9.0 {\small $\pm$ 0.9} & 84.7 {\small $\pm$ 0.5} \\
Qwen 3-32B & 10.5 {\small $\pm$ 0.2} & 1.0 {\small $\pm$ 0.2} & 88.5 {\small $\pm$ 0.1} & 10.6 {\small $\pm$ 0.2} & 10.6 {\small $\pm$ 0.2} & 81.9 {\small $\pm$ 0.6} \\
\bottomrule
\end{tabular}%
}
\vspace{-0.2cm}
\caption{Results of different models on KoSimpleQA. All metrics are reported on a scale of 0 to 100 for better readability. Values represent the mean of three seeds, and $\pm$ denotes the standard deviation. Models are grouped by family and listed in alphabetical order. $\downarrow$ indicates lower-is-better, and $\uparrow$ indicates higher-is-better. Not Attempted (NA) is reported without an arrow as it is not inherently directional.}
\vspace{-0.3cm}
\label{tab:main_full}
\end{table*}

\vspace{-0.2cm}
\section{Experimental Setup}
\vspace{-0.2cm}
To assess model behavior on KoSimpleQA, we evaluate a diverse set of baseline models. We describe the experimental setup below.

\vspace{-0.2cm}
\subsection{Baselines}
\vspace{-0.2cm}
Our evaluation encompasses a diverse suite of open-source models with fewer than 70 billion parameters, specifically selected for their proficiency in the Korean language. We categorize these models into two distinct groups: \textit{Korean Community LLMs} and \textit{Multilingual LLMs}. The Korean Community models include EXAONE 4.0~\citep{research2025exaone}, HyperCLOVA X (HCX) SEED~\citep{team2025hyperclova}, Kanana 1.5~\citep{bak2025kanana}, and VARCO.\footnote{\url{https://huggingface.co/NCSOFT/Llama-VARCO-8B-Instruct}} The Multilingual models supporting Korean comprise Gemma 3~\citep{team2025gemma}, Llama 3.1~\citep{dubey2024llama}, and Qwen3~\citep{yang2025qwen3}.

\vspace{-0.2cm}
\subsection{Implementation Details}
\vspace{-0.2cm}
We adhered to the sampling hyperparameters specified in the SimpleQA framework\footnote{\url{https://github.com/openai/simple-evals}}, setting the temperature to $1.0$ and \texttt{top\_p} to $1.0$, while extending the maximum generation length to 8192 tokens to support reasoning models.
All reported metrics represent the average across three runs with different random seeds. 
For ECE calculation, we used a bin size of $10$.

\begin{figure}[t!]
    \centering
    \begin{minipage}{0.48\textwidth}
        \centering
        \includegraphics[width=\linewidth]{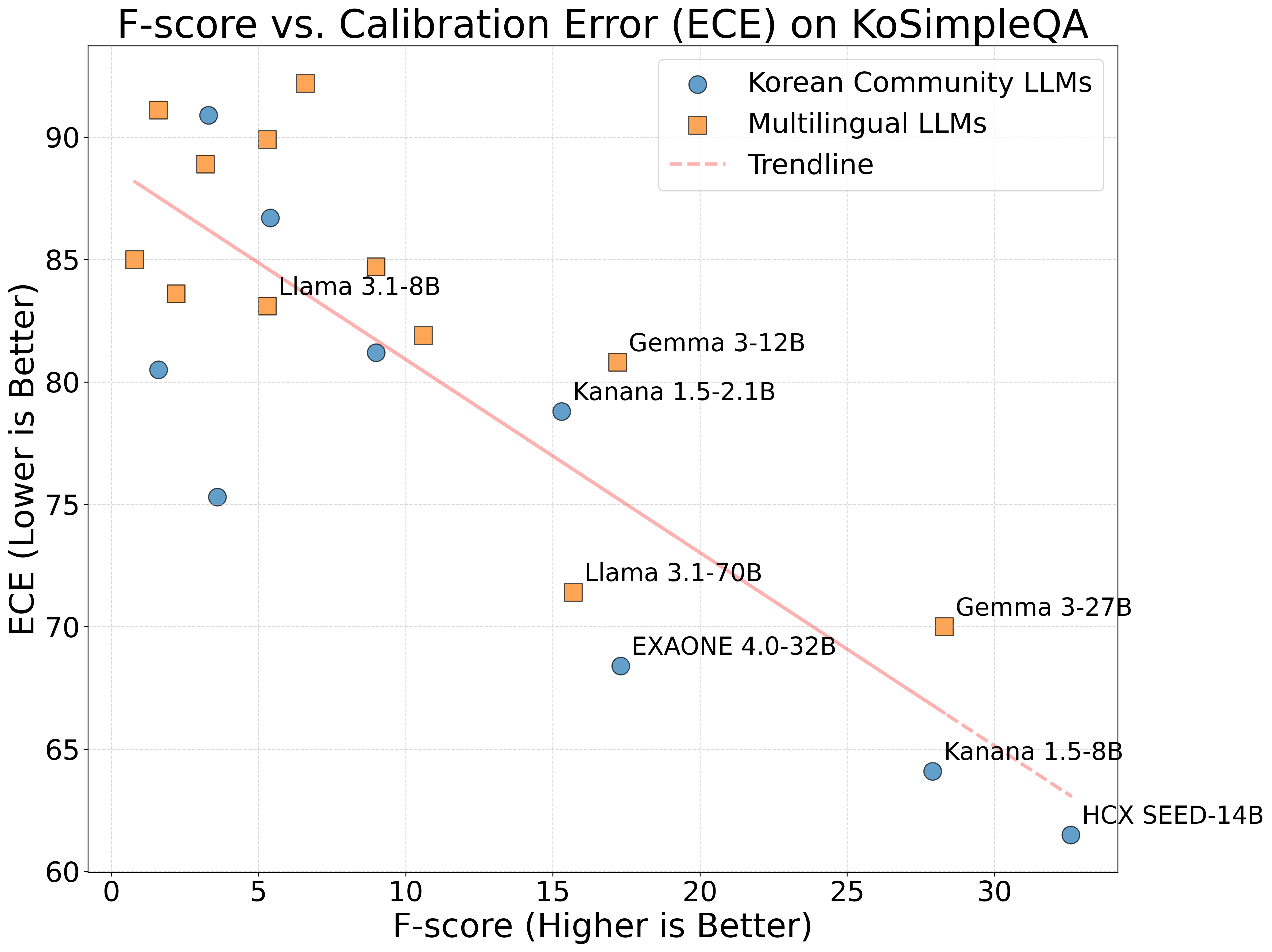}    
        \vspace{-0.2cm}
        \caption{The relationship between performance (F-score) and calibration (ECE) across models on KoSimpleQA.}
        \label{fig:fscore_vs_ece}            
    \end{minipage}
    \hfill 
    \begin{minipage}{0.48\textwidth}
        \centering
        \includegraphics[width=\linewidth]{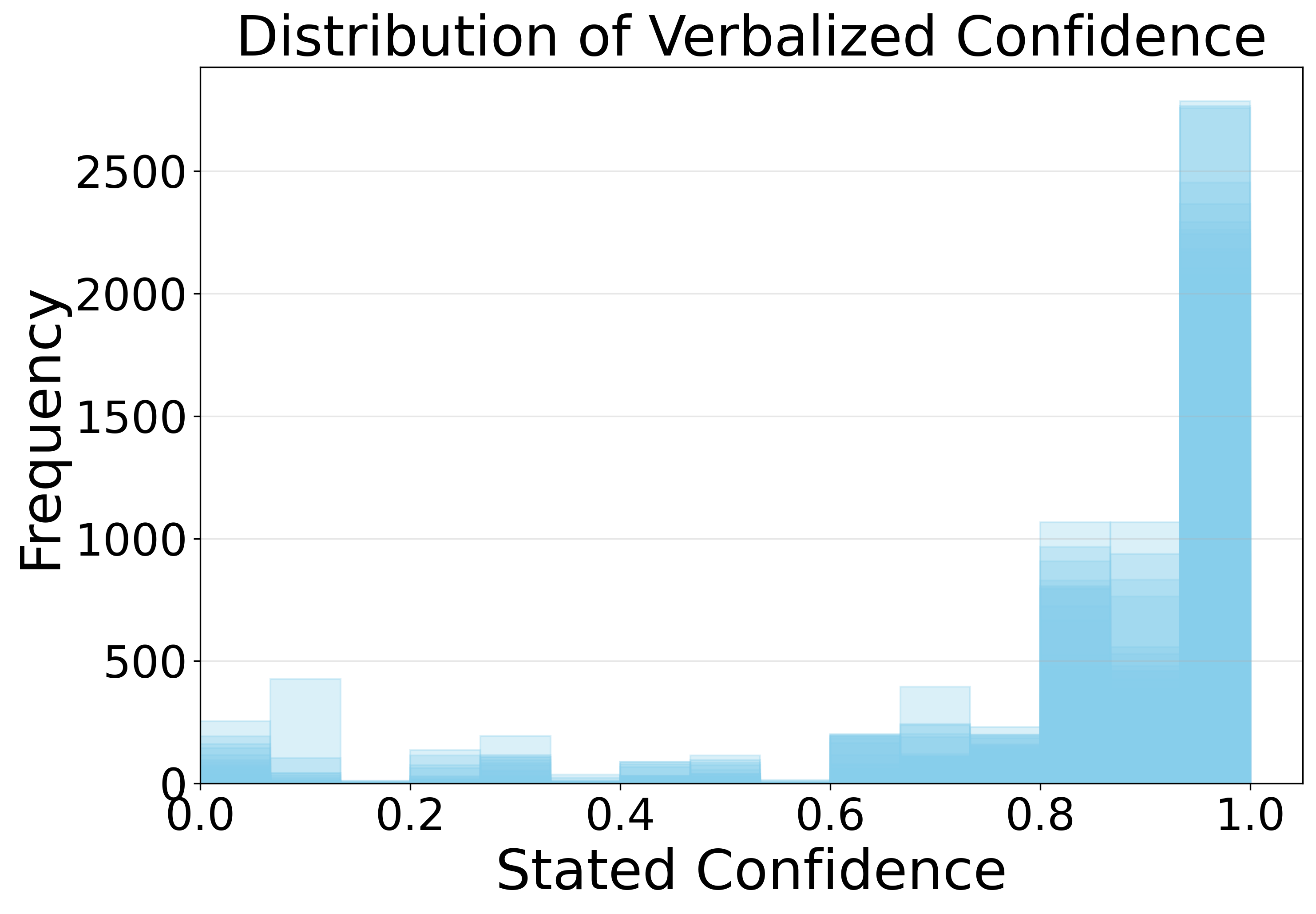}    
        \vspace{-0.2cm}
        \caption{Distribution of stated confidence scores across all evaluated models. The confidence scores are uniformly binned into 15 intervals. }
        \label{fig:confidence_hist}
    \end{minipage}
    \vspace{-0.3cm}
\end{figure}

\vspace{-0.2cm}
\section{Main Results}
\vspace{-0.2cm}
Table \ref{tab:main_full} presents the overall performance of baseline models on KoSimpleQA.
Our analysis highlights three key observations regarding scaling behavior, role of pre-training data distribution and calibration.

\vspace{-0.2cm}
\subsection{Scaling Behavior}
Consistent with established scaling laws \citep{kaplan2020scaling}, performance monotonically improves with model size within each model family. For instance, scaling the HCX SEED family from 0.5B to 14B results in a dramatic F-score increase from 3.6 to 32.6. This trend is pervasive across both Korean and multilingual families, confirming that parametric capacity remains a foundational factor for encoding and accessing factual knowledge.

\vspace{-0.2cm}
\subsection{Role of Pre-training Data Distribution}
While performance generally scales with model size, our results suggest taht the composition of pre-training data may play a more decisive role than parameter count. Korean-native models consistently outperform significantly larger multilingual counterparts. Notably, HCX SEED-14B achieves the highest F-score (32.6), doubling the performance of Llama 3.1-70B (15.7) despite being five times smaller. Similarly, the compact Kanana 1.5-2.1B (15.3) exceeds the massive Qwen 3-32B (10.6). Even Gemma 3-27B, the strongest multilingual baseline, lags behind smaller Korean models.

Although exact training data distributions are rarely disclosed by model developers, Korean community LLMs are generally reported to focus primarily on Korean and English during development. Therefore, a plausible hypothesis for their performance advantage is that their pre-training corpora contain a higher proportion of Korean data than general multilingual LLMs. Following prior observations such as those discussed by \citet{chang2024large}, differences in pretraining corpus composition may influence the factual knowledge that models can effectively express. These findings suggest that high-quality, region-specific pre-training is a critical determinant of factual performance, often outweighing the benefits of generic scaling. This points to targeted data curation and localized LLM development as a promising direction for better capturing culturally grounded knowledge in non-English, low-resource language settings.

\vspace{-0.2cm}
\subsection{Persistence of Overconfidence}
\label{subsec:confidence_calibration}

As shown in Table~\ref{tab:main_full}, higher factual performance generally corresponds to improved $ECE$ values on KoSimpleQA. This relationship is further evidenced by Figure~\ref{fig:fscore_vs_ece}, which illustrates a strong correlation between a model's F-score and its calibration. 
Notably, the calibration trends align with the performance patterns previously observed regarding model scaling and cultural grounding.
For example, culturally specialized models, such as HCX SEED-14B ($ECE$ 61.5) and Kanana 1.5-8B ($ECE$ 64.1), exhibit significantly lower calibration error than larger models like Llama 3.1-70B ($ECE$ 71.4). 
To investigate the specific nature of these calibration failures, we examine the distribution of verbalized confidence in Figure~\ref{fig:confidence_hist}. The results reveal a systemic tendency toward overconfidence across the dataset; even when generating incorrect responses, models frequently assign a confidence level near 100\%.

These findings highlight the need for overconfidence mitigation in culturally specialized contexts. Although our primary contribution lies in establishing reproducible benchmark analyses rather than introducing a new calibration algorithm, contextualizing these calibration failures paves the way for applying existing mitigation techniques~\citep{wen2024mitigating} and serves as a foundational benchmark for future research in model calibration.

\vspace{-0.2cm}
\subsection{Comparison Between SimpleQA and KoSimpleQA}\label{sec:comparison}
\vspace{-0.2cm}

\begin{figure}[t!]
    \centering
    \begin{minipage}{0.48\textwidth}
        \centering   
        \includegraphics[width=\linewidth]{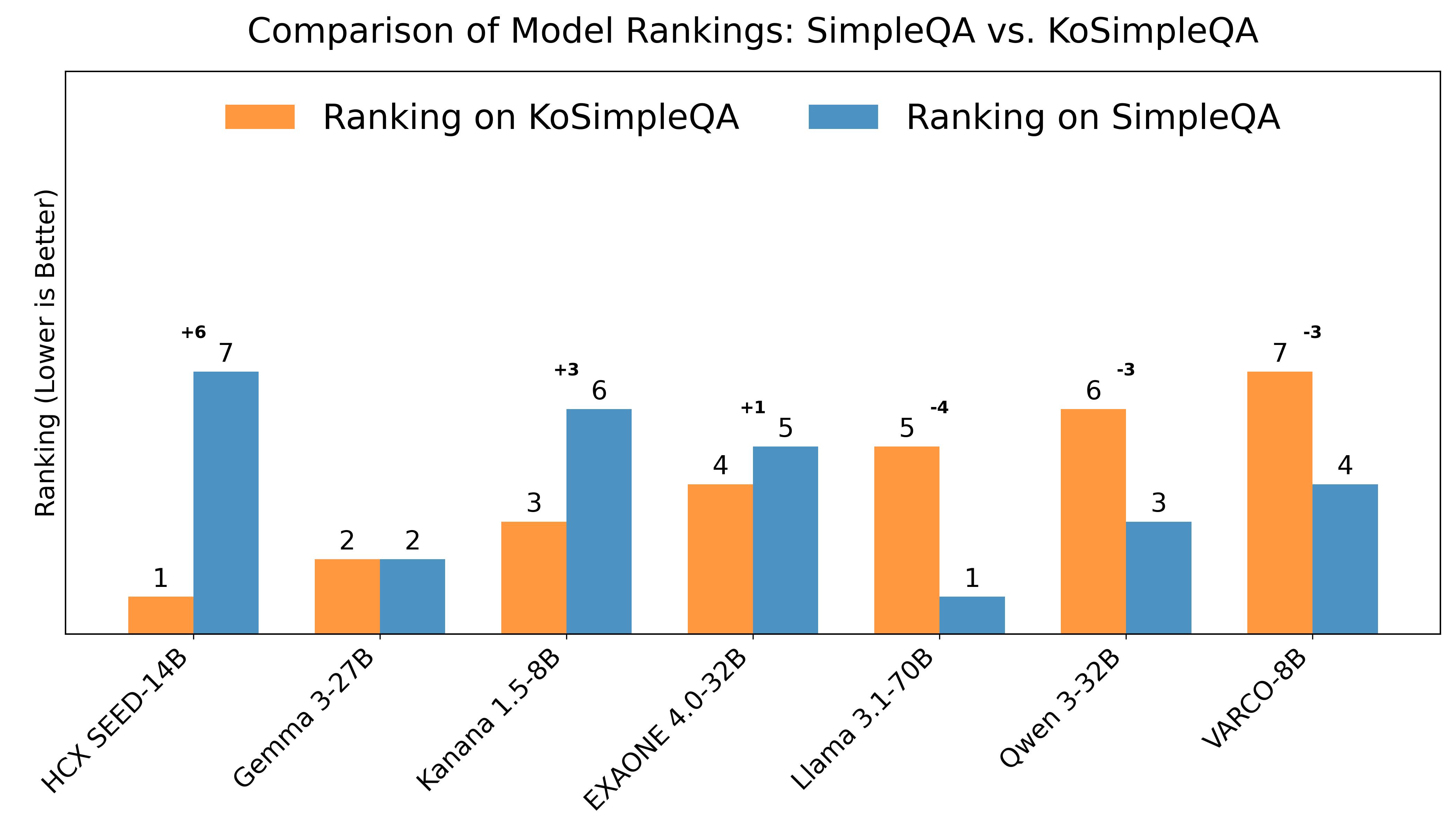}
        \vspace{-0.2cm}
        \caption{Rankings of the largest models in each LLM family on SimpleQA and KoSimpleQA.}
        \label{fig:vs}
    \end{minipage}
    \hfill 
    \begin{minipage}{0.48\textwidth}
        \centering
        \includegraphics[width=\linewidth]{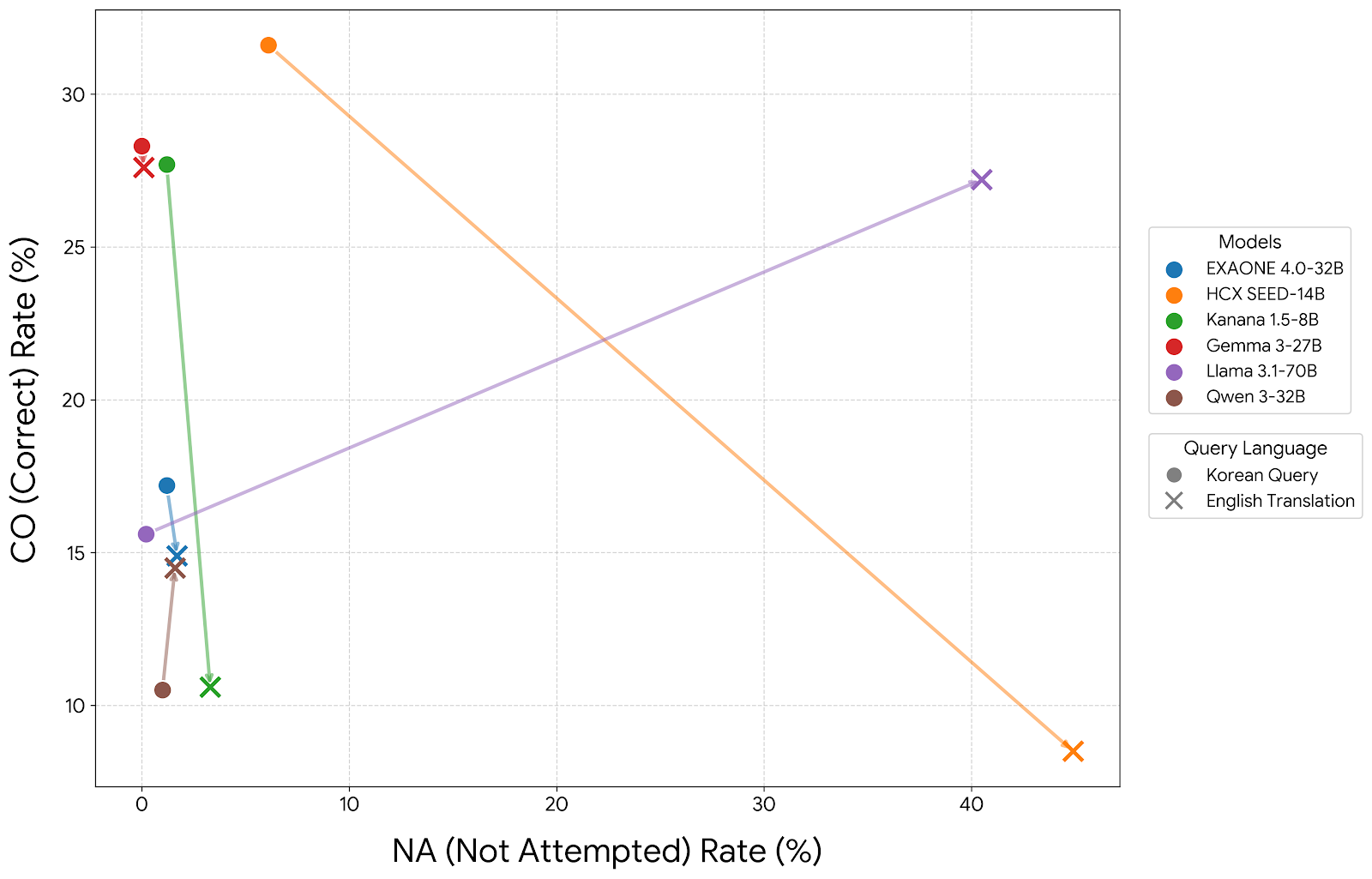}
        \vspace{-0.2cm}
        \caption{
            Impact of query language on model behavior. Arrows indicate the behavioral shift from Korean queries (circles) to English translations (crosses).}
        \label{fig:translate}      
    \end{minipage}
    \vspace{-0.3cm}
\end{figure}

In Figure \ref{fig:vs}, we compare model rankings on English-centric SimpleQA versus our KoSimpleQA to assess cross-lingual factual alignment. We observed a notable \textit{rank reversal}: Llama 3.1-70B drops from \#1 on SimpleQA to \#6 on KoSimpleQA, while HCX SEED-14B jumps from \#8 to \#1. This dramatic shift indicates that high performance on benchmarks reflecting knowledge mainly from one culture may not generalize to another culture. KoSimpleQA thus serves as a critical complement to existing evaluations, exposing the blind spots of high-resource benchmarks and properly validating models tailored for specific linguistic communities. Full results on SimpleQA are provided in Appendix \ref{app:original_simpleqa}.



\vspace{-0.2cm}
\subsection{Same Knowledge, Different Behavior}
\label{sec:translated_behavior}
\vspace{-0.2cm}

While comparing the results on SimpleQA and KoSimpleQA, we observed substantial changes in metrics such as NA for certain models. To investigate whether these differences stem from language variation, we conduct an additional experiment. Specifically, we automatically translate the questions and answers of KoSimpleQA into English and produce a translated KoSimpleQA through a single round of human verification. As shown in Figure~\ref{fig:translate}, even though only the language of the questions was changed, the models exhibit distinct behavioral shifts.

HCX SEED-14B shows a substantial decrease in CO accompanied by a large increase in NA, whereas Llama 3.1-70B exhibits pronounced increases in both CO and NA. The marked increase in NA is consistent with the observation reported by \citet{ul-islam-etal-2025-much} that the degree of hallucination can vary across languages.

Notably, CO also changes significantly, confirming that a model's ability to express internal knowledge can vary depending on the query language \citep{aggarwal2025languagemodelsfactualitydepends}. In this work, we use the term \textbf{cross-lingual knowledge gap} to refer strictly to these observed behavioral disparities in knowledge manifestation across languages, rather than to claim a single underlying cause. Such disparities may stem from multiple factors, including language-specific instruction tuning, refusal alignment, decoding behavior, or pre-training distributions.

\begin{table*}[t!]
\centering
\resizebox{0.9\textwidth}{!}{%
\begin{tabular}{l|ccccc|c}
\toprule
\textbf{Models} & \textbf{CO} $\uparrow$ & \textbf{NA} & \textbf{IN} $\downarrow$ & \textbf{CGA} $\uparrow$ & \textbf{F-score} $\uparrow$ & \textbf{ECE} $\downarrow$ \\
\midrule\midrule
\multicolumn{7}{c}{\textit{Korean Community LLMs}} \\
\midrule
EXAONE 4.0-1.2B-Think & 1.4 {\scriptsize (\color{red}-0.2)} & 1.2 {\scriptsize (\color{red}-3.5)} & 97.4 {\scriptsize (\color{red}+3.7)} & 1.4 {\scriptsize (\color{red}-0.3)} & 1.4 {\scriptsize (\color{red}-0.2)} & 81.6 {\scriptsize (\color{red}+1.1)} \\
EXAONE 4.0-32B-Think & 16.1 {\scriptsize (\color{red}-1.1)} & 0.9 {\scriptsize (\color{red}-0.3)} & 83.0 {\scriptsize (\color{red}+1.4)} & 16.3 {\scriptsize (\color{red}-1.1)} & 16.2 {\scriptsize (\color{red}-1.1)} & 77.4 {\scriptsize (\color{red}+9.0)} \\
\midrule
HCX SEED-14B-Think & \textbf{34.0} {\scriptsize (\color{blue}+2.4)} & 1.6 {\scriptsize (\color{red}-4.5)} & \textbf{64.4} {\scriptsize (\color{red}+2.1)} & \textbf{34.5} {\scriptsize (\color{blue}+0.9)} & \textbf{34.3} {\scriptsize (\color{blue}+1.7)} & \textbf{55.8} {\scriptsize (\color{blue}-5.7)} \\
\midrule
\multicolumn{7}{c}{\textit{Multilingual LLMs Supporting Korean}} \\
\midrule
Qwen 3-0.6B-Think & 0.8 {\scriptsize (0.0)} & 1.4 {\scriptsize (\color{blue}+0.7)} & 97.8 {\scriptsize (\color{blue}-0.6)} & 0.8 {\scriptsize (0.0)} & 0.8 {\scriptsize (0.0)} & 81.1 {\scriptsize (\color{blue}-3.9)} \\
Qwen 3-1.7B-Think & 2.4 {\scriptsize (\color{blue}+0.8)} & 1.7 {\scriptsize (\color{blue}+1.0)} & 95.9 {\scriptsize (\color{blue}-1.8)} & 2.5 {\scriptsize (\color{blue}+0.9)} & 2.4 {\scriptsize (\color{blue}+0.8)} & 84.1 {\scriptsize (\color{blue}-7.0)} \\
Qwen 3-4B-Think & 4.9 {\scriptsize (\color{blue}+1.7)} & 1.6 {\scriptsize (\color{blue}+0.6)} & 93.5 {\scriptsize (\color{blue}-2.4)} & 5.0 {\scriptsize (\color{blue}+1.8)} & 4.9 {\scriptsize (\color{blue}+1.7)} & 82.0 {\scriptsize (\color{blue}-6.9)} \\
Qwen 3-8B-Think & 7.8 {\scriptsize (\color{blue}+2.5)} & 2.9 {\scriptsize (\color{blue}+2.2)} & 89.2 {\scriptsize (\color{blue}-4.8)} & 8.1 {\scriptsize (\color{blue}+2.8)} & 7.9 {\scriptsize (\color{blue}+2.6)} & 76.6 {\scriptsize (\color{blue}-13.3)} \\
Qwen 3-14B-Think & 12.0 {\scriptsize (\color{blue}+3.0)} & 2.4 {\scriptsize (\color{blue}+1.3)} & 85.6 {\scriptsize (\color{blue}-4.3)} & 12.3 {\scriptsize (\color{blue}+3.2)} & 12.1 {\scriptsize (\color{blue}+3.1)} & 73.2 {\scriptsize (\color{blue}-11.5)} \\
Qwen 3-32B-Think & 12.5 {\scriptsize (\color{blue}+2.0)} & 1.0 {\scriptsize (0.0)} & 86.5 {\scriptsize (\color{blue}-2.0)} & 12.6 {\scriptsize (\color{blue}+2.0)} & 12.5 {\scriptsize (\color{blue}+1.9)} & 74.6 {\scriptsize (\color{blue}-7.3)} \\
\bottomrule
\end{tabular}%
}
\vspace{-0.2cm}
\caption{Results of thinking models on KoSimpleQA compared to their base versions. Values in parentheses denote the delta ($\Delta$) from the base model performance (\blueh{blue}: increase, \redh{red}: decrease). Best scores are in \textbf{bold}.}
\label{tab:thinking_delta}
\end{table*}

\vspace{-0.2cm}
\section{Analysis of Reasoning LLMs}
\label{sec:reasoning_analysis}
Recent advancements in LLMs have shown that explicit reasoning can significantly enhance performance in complex tasks.
In this section, we analyze how such reasoning capabilities interact with a model's factual knowledge in the context of KoSimpleQA. We find that while reasoning cannot substitute for missing parametric knowledge, it serves as a powerful amplifier that helps models access and verify latent information.

\vspace{-0.2cm}
\subsection{Reasoning LLMs in Factual Knowledge}

Table~\ref{tab:thinking_delta} presents the performance of models utilizing explicit reasoning (denoted as -Think). We observe an improvement in factual accuracy across most model families when reasoning is enabled, contrary to the intuition that reasoning would not be particularly helpful for factual question answering tasks. We further analyze this phonomenon in Section~\ref{sec:eliciting_suppressed_knowledge}. 

For the HCX SEED-14B model, the F-score increases from 32.6 (Base) to 34.3 (Think), achieving the highest accuracy in our benchmark. Beyond raw accuracy, a notable finding is the improvement in calibration. With reasoning enabled, HCX SEED-14B-Think reaches an $ECE$ of 55.8. This suggests that the internal "thinking" process allows the model to better self-evaluate the correctness of its retrieved facts. Nevertheless, for models with lower parametric capacity, such as the Qwen 3-0.6B and 1.7B variants, reasoning provides negligible gains or even slightly inflates $ECE$.

\vspace{-0.2cm}
\subsection{Eliciting Suppressed Knowledge}\label{sec:eliciting_suppressed_knowledge}

Section~\ref{sec:comparison} demonstrates that a model’s knowledge may not be fully expressed depending on the language of the query. Table~\ref{tab:kosimpleqa_co_na} further illustrates how reasoning interacts with this phenomenon. Specifically, we observe that reasoning yields substantially larger gains in a model’s weaker language, while providing only marginal improvements in its dominant language. For instance, HCX SEED-14B, which is stronger in Korean, shows a modest gain of 2.4 in Korean but a significantly larger gain of 6.8 in English. Conversely, Qwen 3-32B, which is stronger in English, exhibits a gain of 0.4 in English but 2.0 in Korean.

These results suggest that reasoning does not primarily enhance factual question answering performance, but rather serves to mitigate cross-lingual knowledge gaps. This observation aligns with our intuition that, for tasks such as KoSimpleQA that emphasize knowledge memorization rather than reasoning, explicit reasoning provides limited additional benefit when the underlying knowledge is already accessible. Qualitative examples illustrating how reasoning can elicit latent knowledge are provided in Appendix~\ref{subsec:qualitative_reasoning} (Table~\ref{tab:base_vs_think_english_first}).

Interestingly, Exaone does not exhibit substantial improvements with reasoning. We hypothesize that this discrepancy arises from differences in training methodologies for reasoning capabilities. This suggests that while reasoning has the potential to alleviate cross-lingual knowledge gaps, its effectiveness is contingent on how such capabilities are learned and integrated during training.

Another interesting finding is that reasoning substantially reduces NA rates, particularly in settings where models exhibit excessive refusal behavior. For example, HCX SEED-14B shows a reduction of 44.3 in NA on English data. This observation is consistent with prior work~\citep{yin2025refusalfallscliffsafety}, which reports that enabling reasoning can lead to decreased refusal behavior.




\begin{table}[t!]
\centering
\resizebox{0.6\columnwidth}{!}{%
\begin{tabular}{lc|cc|cc}
\toprule
\multirow{2}{*}{\textbf{Model}} & \multirow{2}{*}{\textbf{Metric}} & \multicolumn{2}{c|}{\textbf{Korean Prompt}} & \multicolumn{2}{c}{\textbf{Translated (English)}} \\ 
\cmidrule(lr){3-4} \cmidrule(lr){5-6}
 & & Base &  Think & Base & Think \\ 
\midrule\midrule

\multirow{2}{*}{EXAONE 4.0-32B} & CO & 17.2 & 16.1 & 14.9 & 15.3 \\ 
 & NA & 1.2 & 0.9 & 1.7 & 1.6 \\ 
\midrule

\multirow{2}{*}{HCX SEED-14B} & CO & 31.6 & 34.0 & 8.5 & 15.3 \\ 
 & NA & 6.1 & 1.6 & 44.9 & 0.6 \\ 
\midrule

\multirow{2}{*}{Qwen 3-32B} & CO & 10.5 & 12.5 & 14.5 & 14.9 \\ 
 & NA & 1.0 & 1.0 & 1.6 & 2.1 \\ 
\bottomrule
\end{tabular}%
}
\vspace{-0.2cm}
\caption{Effect of Reasoning in the Translation Experiment: Comparison of CO and NA.}
\label{tab:kosimpleqa_co_na}
\vspace{-0.2cm}
\end{table}

\section{Conclusion}

In this work, we introduce KoSimpleQA, a benchmark designed to evaluate short-form factuality within Korean cultural contexts. Our evaluation across diverse LLMs reveals that current models still struggle with factual accuracy in Korean, exhibiting performance rankings that diverge significantly from English-centric benchmarks. These findings highlight both the substantial room for improvement in non-English, low-resource language settings and the importance of localized, culturally grounded evaluation. Furthermore, our analysis of reasoning models indicates its potential to mitigate cross-lingual knowledge gap. We expect KoSimpleQA to serve as a vital foundation for developing more culturally attuned and factually reliable language models.

\bibliography{colm2026_conference}

@article{wei2024measuring,
  title={Measuring short-form factuality in large language models},
  author={Wei, Jason and Karina, Nguyen and Chung, Hyung Won and Jiao, Yunxin Joy and Papay, Spencer and Glaese, Amelia and Schulman, John and Fedus, William},
  journal={arXiv preprint arXiv:2411.04368},
  year={2024}
}

@article{he2024chinese,
  title={Chinese simpleqa: A chinese factuality evaluation for large language models},
  author={He, Yancheng and Li, Shilong and Liu, Jiaheng and Tan, Yingshui and Wang, Weixun and Huang, Hui and Bu, Xingyuan and Guo, Hangyu and Hu, Chengwei and Zheng, Boren and others},
  journal={arXiv preprint arXiv:2411.07140},
  year={2024}
}

@inproceedings{banerjee2025llms,
  title={Llms will always hallucinate, and we need to live with this},
  author={Banerjee, Sourav and Agarwal, Ayushi and Singla, Saloni},
  booktitle={Intelligent Systems Conference},
  pages={624--648},
  year={2025},
  organization={Springer}
}

@article{yoo2024hyperclova,
  title={Hyperclova x technical report},
  author={Yoo, Kang Min and Han, Jaegeun and In, Sookyo and Jeon, Heewon and Jeong, Jisu and Kang, Jaewook and Kim, Hyunwook and Kim, Kyung-Min and Kim, Munhyong and Kim, Sungju and others},
  journal={arXiv preprint arXiv:2404.01954},
  year={2024}
}

@article{hurst2024gpt,
  title={Gpt-4o system card},
  author={Hurst, Aaron and Lerer, Adam and Goucher, Adam P and Perelman, Adam and Ramesh, Aditya and Clark, Aidan and Ostrow, AJ and Welihinda, Akila and Hayes, Alan and Radford, Alec and others},
  journal={arXiv preprint arXiv:2410.21276},
  year={2024}
}

@misc{team2025gemma,
      title={Gemma 3 Technical Report}, 
      author={{Gemma Team} and Aishwarya Kamath and Johan Ferret and Shreya Pathak and Nino Vieillard and Ramona Merhej and Sarah Perrin and Tatiana Matejovicova and Alexandre Ramé and Morgane Rivière and Louis Rouillard and Thomas Mesnard and Geoffrey Cideron and Jean-bastien Grill and Sabela Ramos and Edouard Yvinec and Michelle Casbon and Etienne Pot and Ivo Penchev and Gaël Liu and Francesco Visin and Kathleen Kenealy and Lucas Beyer and Xiaohai Zhai and Anton Tsitsulin and Robert Busa-Fekete and Alex Feng and Noveen Sachdeva and Benjamin Coleman and Yi Gao and Basil Mustafa and Iain Barr and Emilio Parisotto and David Tian and Matan Eyal and Colin Cherry and Jan-Thorsten Peter and Danila Sinopalnikov and Surya Bhupatiraju and Rishabh Agarwal and Mehran Kazemi and Dan Malkin and Ravin Kumar and David Vilar and Idan Brusilovsky and Jiaming Luo and Andreas Steiner and Abe Friesen and Abhanshu Sharma and Abheesht Sharma and Adi Mayrav Gilady and Adrian Goedeckemeyer and Alaa Saade and Alex Feng and Alexander Kolesnikov and Alexei Bendebury and Alvin Abdagic and Amit Vadi and András György and André Susano Pinto and Anil Das and Ankur Bapna and Antoine Miech and Antoine Yang and Antonia Paterson and Ashish Shenoy and Ayan Chakrabarti and Bilal Piot and Bo Wu and Bobak Shahriari and Bryce Petrini and Charlie Chen and Charline Le Lan and Christopher A. Choquette-Choo and CJ Carey and Cormac Brick and Daniel Deutsch and Danielle Eisenbud and Dee Cattle and Derek Cheng and Dimitris Paparas and Divyashree Shivakumar Sreepathihalli and Doug Reid and Dustin Tran and Dustin Zelle and Eric Noland and Erwin Huizenga and Eugene Kharitonov and Frederick Liu and Gagik Amirkhanyan and Glenn Cameron and Hadi Hashemi and Hanna Klimczak-Plucińska and Harman Singh and Harsh Mehta and Harshal Tushar Lehri and Hussein Hazimeh and Ian Ballantyne and Idan Szpektor and Ivan Nardini and Jean Pouget-Abadie and Jetha Chan and Joe Stanton and John Wieting and Jonathan Lai and Jordi Orbay and Joseph Fernandez and Josh Newlan and Ju-yeong Ji and Jyotinder Singh and Kat Black and Kathy Yu and Kevin Hui and Kiran Vodrahalli and Klaus Greff and Linhai Qiu and Marcella Valentine and Marina Coelho and Marvin Ritter and Matt Hoffman and Matthew Watson and Mayank Chaturvedi and Michael Moynihan and Min Ma and Nabila Babar and Natasha Noy and Nathan Byrd and Nick Roy and Nikola Momchev and Nilay Chauhan and Noveen Sachdeva and Oskar Bunyan and Pankil Botarda and Paul Caron and Paul Kishan Rubenstein and Phil Culliton and Philipp Schmid and Pier Giuseppe Sessa and Pingmei Xu and Piotr Stanczyk and Pouya Tafti and Rakesh Shivanna and Renjie Wu and Renke Pan and Reza Rokni and Rob Willoughby and Rohith Vallu and Ryan Mullins and Sammy Jerome and Sara Smoot and Sertan Girgin and Shariq Iqbal and Shashir Reddy and Shruti Sheth and Siim Põder and Sijal Bhatnagar and Sindhu Raghuram Panyam and Sivan Eiger and Susan Zhang and Tianqi Liu and Trevor Yacovone and Tyler Liechty and Uday Kalra and Utku Evci and Vedant Misra and Vincent Roseberry and Vlad Feinberg and Vlad Kolesnikov and Woohyun Han and Woosuk Kwon and Xi Chen and Yinlam Chow and Yuvein Zhu and Zichuan Wei and Zoltan Egyed and Victor Cotruta and Minh Giang and Phoebe Kirk and Anand Rao and Kat Black and Nabila Babar and Jessica Lo and Erica Moreira and Luiz Gustavo Martins and Omar Sanseviero and Lucas Gonzalez and Zach Gleicher and Tris Warkentin and Vahab Mirrokni and Evan Senter and Eli Collins and Joelle Barral and Zoubin Ghahramani and Raia Hadsell and Yossi Matias and D. Sculley and Slav Petrov and Noah Fiedel and Noam Shazeer and Oriol Vinyals and Jeff Dean and Demis Hassabis and Koray Kavukcuoglu and Clement Farabet and Elena Buchatskaya and Jean-Baptiste Alayrac and Rohan Anil and Dmitry and Lepikhin and Sebastian Borgeaud and Olivier Bachem and Armand Joulin and Alek Andreev and Cassidy Hardin and Robert Dadashi and Léonard Hussenot},
      year={2025},
      eprint={2503.19786},
      archivePrefix={arXiv},
      primaryClass={cs.CL},
      url={https://arxiv.org/abs/2503.19786}, 
}

@article{dubey2024llama,
  title={The llama 3 herd of models},
  author={Dubey, Abhimanyu and Jauhri, Abhinav and Pandey, Abhinav and Kadian, Abhishek and Al-Dahle, Ahmad and Letman, Aiesha and Mathur, Akhil and Schelten, Alan and Yang, Amy and Fan, Angela and others},
  journal={arXiv e-prints},
  pages={arXiv--2407},
  year={2024}
}

@article{yang2025qwen3,
  title={Qwen3 technical report},
  author={Yang, An and Li, Anfeng and Yang, Baosong and Zhang, Beichen and Hui, Binyuan and Zheng, Bo and Yu, Bowen and Gao, Chang and Huang, Chengen and Lv, Chenxu and others},
  journal={arXiv preprint arXiv:2505.09388},
  year={2025}
}

@article{kaplan2020scaling,
  title={Scaling laws for neural language models},
  author={Kaplan, Jared and McCandlish, Sam and Henighan, Tom and Brown, Tom B and Chess, Benjamin and Child, Rewon and Gray, Scott and Radford, Alec and Wu, Jeffrey and Amodei, Dario},
  journal={arXiv preprint arXiv:2001.08361},
  year={2020}
}

@misc{research2025exaone,
      title={EXAONE Deep: Reasoning Enhanced Language Models}, 
      author={{LG AI Research} and Kyunghoon Bae and Eunbi Choi and Kibong Choi and Stanley Jungkyu Choi and Yemuk Choi and Seokhee Hong and Junwon Hwang and Hyojin Jeon and Kijeong Jeon and Gerrard Jeongwon Jo and Hyunjik Jo and Jiyeon Jung and Hyosang Kim and Joonkee Kim and Seonghwan Kim and Soyeon Kim and Sunkyoung Kim and Yireun Kim and Yongil Kim and Youchul Kim and Edward Hwayoung Lee and Haeju Lee and Honglak Lee and Jinsik Lee and Kyungmin Lee and Sangha Park and Yongmin Park and Sihoon Yang and Heuiyeen Yeen and Sihyuk Yi and Hyeongu Yun},
      year={2025},
      eprint={2503.12524},
      archivePrefix={arXiv},
      primaryClass={cs.CL},
      url={https://arxiv.org/abs/2503.12524}, 
}

@misc{haas2025verified,
      title={SimpleQA Verified: A Reliable Factuality Benchmark to Measure Parametric Knowledge}, 
      author={Lukas Haas and Gal Yona and Giovanni D'Antonio and Sasha Goldshtein and Dipanjan Das},
      year={2025},
      eprint={2509.07968},
      archivePrefix={arXiv},
      primaryClass={cs.CL},
      url={https://arxiv.org/abs/2509.07968}, 
}

@inproceedings{son2024haerae,
  title={Hae-rae bench: Evaluation of korean knowledge in language models},
  author={Son, Guijin and Lee, Hanwool and Kim, Suwan and Kim, Huiseo and cheol Lee, Jae and Yeom, Je Won and Jung, Jihyu and woo Kim, Jung and Kim, Songseong},
  booktitle={Proceedings of the 2024 Joint International Conference on Computational Linguistics, Language Resources and Evaluation (LREC-COLING 2024)},
  pages={7993--8007},
  year={2024}
}

@inproceedings{wang2024kulture,
    title = "{KULTURE} Bench: A Benchmark for Assessing Language Model in {K}orean Cultural Context",
    author = "Wang, Xiaonan  and
      Yeo, Jinyoung  and
      Lim, Joon-Ho  and
      Kim, Hansaem",
    editor = "Oco, Nathaniel  and
      Dita, Shirley N.  and
      Borlongan, Ariane Macalinga  and
      Kim, Jong-Bok",
    booktitle = "Proceedings of the 38th Pacific Asia Conference on Language, Information and Computation",
    month = dec,
    year = "2024",
    address = "Tokyo, Japan",
    publisher = "Tokyo University of Foreign Studies",
    url = "https://aclanthology.org/2024.paclic-1.88/",
    pages = "914--927"
}

@inproceedings{kim2024click,
  title={CLIcK: A Benchmark Dataset of Cultural and Linguistic Intelligence in Korean},
  author={Kim, Eunsu and Suk, Juyoung and Oh, Philhoon and Yoo, Haneul and Thorne, James and Oh, Alice},
  booktitle={Proceedings of the 2024 Joint International Conference on Computational Linguistics, Language Resources and Evaluation (LREC-COLING 2024)},
  pages={3335--3346},
  year={2024}
}

@inproceedings{lee-etal-2024-kornat,
    title = "{K}or{NAT}: {LLM} Alignment Benchmark for {K}orean Social Values and Common Knowledge",
    author = "Lee, Jiyoung  and
      Kim, Minwoo  and
      Kim, Seungho  and
      Kim, Junghwan  and
      Won, Seunghyun  and
      Lee, Hwaran  and
      Choi, Edward",
    editor = "Ku, Lun-Wei  and
      Martins, Andre  and
      Srikumar, Vivek",
    booktitle = "Findings of the Association for Computational Linguistics: ACL 2024",
    month = aug,
    year = "2024",
    address = "Bangkok, Thailand",
    publisher = "Association for Computational Linguistics",
    url = "https://aclanthology.org/2024.findings-acl.666/",
    doi = "10.18653/v1/2024.findings-acl.666",
    pages = "11177--11213"
}

@inproceedings{
    xiong2024can,
    title={Can {LLM}s Express Their Uncertainty? An Empirical Evaluation of Confidence Elicitation in {LLM}s},
    author={Miao Xiong and Zhiyuan Hu and Xinyang Lu and YIFEI LI and Jie Fu and Junxian He and Bryan Hooi},
    booktitle={The Twelfth International Conference on Learning Representations},
    year={2024},
    url={https://openreview.net/forum?id=gjeQKFxFpZ}
}

@inproceedings{ece2015,
    author = {Naeini, Mahdi Pakdaman and Cooper, Gregory F. and Hauskrecht, Milos},
    title = {Obtaining well calibrated probabilities using bayesian binning},
    year = {2015},
    isbn = {0262511290},
    publisher = {AAAI Press},
    booktitle = {Proceedings of the Twenty-Ninth AAAI Conference on Artificial Intelligence},
    pages = {2901–2907},
    numpages = {7},
    location = {Austin, Texas},
    series = {AAAI'15}
}

@inproceedings{lin2022truthfulqa,
    title = "{T}ruthful{QA}: Measuring How Models Mimic Human Falsehoods",
    author = "Lin, Stephanie  and
      Hilton, Jacob  and
      Evans, Owain",
    editor = "Muresan, Smaranda  and
      Nakov, Preslav  and
      Villavicencio, Aline",
    booktitle = "Proceedings of the 60th Annual Meeting of the Association for Computational Linguistics (Volume 1: Long Papers)",
    month = may,
    year = "2022",
    address = "Dublin, Ireland",
    publisher = "Association for Computational Linguistics",
    url = "https://aclanthology.org/2022.acl-long.229/",
    doi = "10.18653/v1/2022.acl-long.229",
    pages = "3214--3252"
}

@article{bak2025kanana,
  title={Kanana: Compute-efficient bilingual language models},
  author={Bak, Yunju and Lee, Hojin and Ryu, Minho and Ham, Jiyeon and Jung, Seungjae and Nam, Daniel Wontae and Eo, Taegyeong and Lee, Donghun and Jung, Doohae and Kim, Boseop and others},
  journal={arXiv preprint arXiv:2502.18934},
  year={2025}
}

@article{team2025hyperclova,
  title={HyperCLOVA X THINK Technical Report},
  author={Team, NAVER Cloud HyperCLOVA X},
  journal={arXiv preprint arXiv:2506.22403},
  year={2025}
}

@inproceedings{
    yoon2025reasoning,
    title={Reasoning Models Better Express Their Confidence},
    author={Dongkeun Yoon and Seungone Kim and Sohee Yang and Sunkyoung Kim and Soyeon Kim and Yongil Kim and Eunbi Choi and Yireun Kim and Minjoon Seo},
    booktitle={The Thirty-ninth Annual Conference on Neural Information Processing Systems},
    year={2025},
    url={https://openreview.net/forum?id=rbBtoVnduo}
}

@inproceedings{ul-islam-etal-2025-much,
    title = "How Much Do {LLM}s Hallucinate across Languages? On Realistic Multilingual Estimation of {LLM} Hallucination",
    author = "Ul Islam, Saad Obaid  and
      Lauscher, Anne  and
      Glava{\v{s}}, Goran",
    editor = "Christodoulopoulos, Christos  and
      Chakraborty, Tanmoy  and
      Rose, Carolyn  and
      Peng, Violet",
    booktitle = "Proceedings of the 2025 Conference on Empirical Methods in Natural Language Processing",
    month = nov,
    year = "2025",
    address = "Suzhou, China",
    publisher = "Association for Computational Linguistics",
    url = "https://aclanthology.org/2025.emnlp-main.1481/",
    doi = "10.18653/v1/2025.emnlp-main.1481",
    pages = "29077--29098",
    ISBN = "979-8-89176-332-6"
}

@misc{yin2025refusalfallscliffsafety,
      title={Refusal Falls off a Cliff: How Safety Alignment Fails in Reasoning?}, 
      author={Qingyu Yin and Chak Tou Leong and Linyi Yang and Wenxuan Huang and Wenjie Li and Xiting Wang and Jaehong Yoon and YunXing and XingYu and Jinjin Gu},
      year={2025},
      eprint={2510.06036},
      archivePrefix={arXiv},
      primaryClass={cs.AI},
      url={https://arxiv.org/abs/2510.06036}, 
}

@inproceedings{geng-etal-2024-survey,
    title = "A Survey of Confidence Estimation and Calibration in Large Language Models",
    author = "Geng, Jiahui  and
      Cai, Fengyu  and
      Wang, Yuxia  and
      Koeppl, Heinz  and
      Nakov, Preslav  and
      Gurevych, Iryna",
    editor = "Duh, Kevin  and
      Gomez, Helena  and
      Bethard, Steven",
    booktitle = "Proceedings of the 2024 Conference of the North American Chapter of the Association for Computational Linguistics: Human Language Technologies (Volume 1: Long Papers)",
    month = jun,
    year = "2024",
    address = "Mexico City, Mexico",
    publisher = "Association for Computational Linguistics",
    url = "https://aclanthology.org/2024.naacl-long.366/",
    doi = "10.18653/v1/2024.naacl-long.366",
    pages = "6577--6595"
}

@inproceedings{tian-etal-2023-just,
    title = "Just Ask for Calibration: Strategies for Eliciting Calibrated Confidence Scores from Language Models Fine-Tuned with Human Feedback",
    author = "Tian, Katherine  and
      Mitchell, Eric  and
      Zhou, Allan  and
      Sharma, Archit  and
      Rafailov, Rafael  and
      Yao, Huaxiu  and
      Finn, Chelsea  and
      Manning, Christopher",
    editor = "Bouamor, Houda  and
      Pino, Juan  and
      Bali, Kalika",
    booktitle = "Proceedings of the 2023 Conference on Empirical Methods in Natural Language Processing",
    month = dec,
    year = "2023",
    address = "Singapore",
    publisher = "Association for Computational Linguistics",
    url = "https://aclanthology.org/2023.emnlp-main.330/",
    doi = "10.18653/v1/2023.emnlp-main.330",
    pages = "5433--5442"
}

@misc{aggarwal2025languagemodelsfactualitydepends,
      title={Language Models' Factuality Depends on the Language of Inquiry}, 
      author={Tushar Aggarwal and Kumar Tanmay and Ayush Agrawal and Kumar Ayush and Hamid Palangi and Paul Pu Liang},
      year={2025},
      eprint={2502.17955},
      archivePrefix={arXiv},
      primaryClass={cs.CL},
      url={https://arxiv.org/abs/2502.17955}, 
}

@inproceedings{tam-etal-2025-none,
    title = "None of the Above, Less of the Right Parallel Patterns in Human and {LLM} Performance on Multi-Choice Questions Answering",
    author = "Tam, Zhi Rui  and
      Wu, Cheng-Kuang  and
      Lin, Chieh-Yen  and
      Chen, Yun-Nung",
    editor = "Che, Wanxiang  and
      Nabende, Joyce  and
      Shutova, Ekaterina  and
      Pilehvar, Mohammad Taher",
    booktitle = "Findings of the Association for Computational Linguistics: ACL 2025",
    month = jul,
    year = "2025",
    address = "Vienna, Austria",
    publisher = "Association for Computational Linguistics",
    url = "https://aclanthology.org/2025.findings-acl.1031/",
    doi = "10.18653/v1/2025.findings-acl.1031",
    pages = "20112--20134",
    ISBN = "979-8-89176-256-5"
}

@misc{singh2025optionspitfallsmultiplechoicequestions,
      title={It is Too Many Options: Pitfalls of Multiple-Choice Questions in Generative AI and Medical Education}, 
      author={Shrutika Singh and Anton Alyakin and Daniel Alexander Alber and Jaden Stryker and Ai Phuong S Tong and Karl Sangwon and Nicolas Goff and Mathew de la Paz and Miguel Hernandez-Rovira and Ki Yun Park and Eric Claude Leuthardt and Eric Karl Oermann},
      year={2025},
      eprint={2503.13508},
      archivePrefix={arXiv},
      primaryClass={cs.CL},
      url={https://arxiv.org/abs/2503.13508}, 
}

@inproceedings{li-etal-2024-multiple,
    title = "Can Multiple-choice Questions Really Be Useful in Detecting the Abilities of {LLM}s?",
    author = "Li, Wangyue  and
      Li, Liangzhi  and
      Xiang, Tong  and
      Liu, Xiao  and
      Deng, Wei  and
      Garcia, Noa",
    editor = "Calzolari, Nicoletta  and
      Kan, Min-Yen  and
      Hoste, Veronique  and
      Lenci, Alessandro  and
      Sakti, Sakriani  and
      Xue, Nianwen",
    booktitle = "Proceedings of the 2024 Joint International Conference on Computational Linguistics, Language Resources and Evaluation (LREC-COLING 2024)",
    month = may,
    year = "2024",
    address = "Torino, Italia",
    publisher = "ELRA and ICCL",
    url = "https://aclanthology.org/2024.lrec-main.251/",
    pages = "2819--2834"
}

@inproceedings{balepur-etal-2025-best,
    title = "Which of These Best Describes Multiple Choice Evaluation with {LLM}s? A) Forced {B}) Flawed {C}) Fixable {D}) All of the Above",
    author = "Balepur, Nishant  and
      Rudinger, Rachel  and
      Boyd-Graber, Jordan Lee",
    editor = "Che, Wanxiang  and
      Nabende, Joyce  and
      Shutova, Ekaterina  and
      Pilehvar, Mohammad Taher",
    booktitle = "Proceedings of the 63rd Annual Meeting of the Association for Computational Linguistics (Volume 1: Long Papers)",
    month = jul,
    year = "2025",
    address = "Vienna, Austria",
    publisher = "Association for Computational Linguistics",
    url = "https://aclanthology.org/2025.acl-long.169/",
    doi = "10.18653/v1/2025.acl-long.169",
    pages = "3394--3418",
    ISBN = "979-8-89176-251-0"
}

@misc{madhusudhan2024llmsknowanswerinvestigating,
      title={Do LLMs Know When to NOT Answer? Investigating Abstention Abilities of Large Language Models}, 
      author={Nishanth Madhusudhan and Sathwik Tejaswi Madhusudhan and Vikas Yadav and Masoud Hashemi},
      year={2024},
      eprint={2407.16221},
      archivePrefix={arXiv},
      primaryClass={cs.CL},
      url={https://arxiv.org/abs/2407.16221}, 
}

@inproceedings{bandarkar-etal-2024-belebele,
    title = "The Belebele Benchmark: a Parallel Reading Comprehension Dataset in 122 Language Variants",
    author = "Bandarkar, Lucas  and
      Liang, Davis  and
      Muller, Benjamin  and
      Artetxe, Mikel  and
      Shukla, Satya Narayan  and
      Husa, Donald  and
      Goyal, Naman  and
      Krishnan, Abhinandan  and
      Zettlemoyer, Luke  and
      Khabsa, Madian",
    editor = "Ku, Lun-Wei  and
      Martins, Andre  and
      Srikumar, Vivek",
    booktitle = "Proceedings of the 62nd Annual Meeting of the Association for Computational Linguistics (Volume 1: Long Papers)",
    month = aug,
    year = "2024",
    address = "Bangkok, Thailand",
    publisher = "Association for Computational Linguistics",
    url = "https://aclanthology.org/2024.acl-long.44/",
    doi = "10.18653/v1/2024.acl-long.44",
    pages = "749--775"
}

@inproceedings{artetxe-etal-2020-cross,
    title = "On the Cross-lingual Transferability of Monolingual Representations",
    author = "Artetxe, Mikel  and
      Ruder, Sebastian  and
      Yogatama, Dani",
    editor = "Jurafsky, Dan  and
      Chai, Joyce  and
      Schluter, Natalie  and
      Tetreault, Joel",
    booktitle = "Proceedings of the 58th Annual Meeting of the Association for Computational Linguistics",
    month = jul,
    year = "2020",
    address = "Online",
    publisher = "Association for Computational Linguistics",
    url = "https://aclanthology.org/2020.acl-main.421/",
    doi = "10.18653/v1/2020.acl-main.421",
    pages = "4623--4637"
}

@inproceedings{talmor-etal-2019-commonsenseqa,
    title = "{C}ommonsense{QA}: A Question Answering Challenge Targeting Commonsense Knowledge",
    author = "Talmor, Alon  and
      Herzig, Jonathan  and
      Lourie, Nicholas  and
      Berant, Jonathan",
    editor = "Burstein, Jill  and
      Doran, Christy  and
      Solorio, Thamar",
    booktitle = "Proceedings of the 2019 Conference of the North {A}merican Chapter of the Association for Computational Linguistics: Human Language Technologies, Volume 1 (Long and Short Papers)",
    month = jun,
    year = "2019",
    address = "Minneapolis, Minnesota",
    publisher = "Association for Computational Linguistics",
    url = "https://aclanthology.org/N19-1421/",
    doi = "10.18653/v1/N19-1421",
    pages = "4149--4158"
}

@article{chang2024large,
  title={How do large language models acquire factual knowledge during pretraining?},
  author={Chang, Hoyeon and Park, Jinho and Ye, Seonghyeon and Yang, Sohee and Seo, Youngkyung and Chang, Du-Seong and Seo, Minjoon},
  journal={Advances in neural information processing systems},
  volume={37},
  pages={60626--60668},
  year={2024}
}

@inproceedings{wen2024mitigating,
  title={Mitigating overconfidence in large language models: A behavioral lens on confidence estimation and calibration},
  author={Wen, Bingbing and Xu, Chenjun and Wolfe, Robert and Wang, Lucy Lu and Howe, Bill and others},
  booktitle={NeurIPS 2024 Workshop on Behavioral Machine Learning},
  year={2024}
}

@article{jiang2025evaluating,
  title={Evaluating large language model with knowledge oriented language specific simple question answering},
  author={Jiang, Bowen and Zhu, Runchuan and Wu, Jiang and Jiang, Zinco and He, Yifan and Gao, Junyuan and Yu, Jia and Min, Rui and Wang, Yinfan and Yang, Haote and others},
  journal={arXiv preprint arXiv:2505.16591},
  year={2025}
}
\bibliographystyle{colm2026_conference}

\newpage
\appendix

\section{Statistics and examples of KoSimpleQA}
\label{app:stat}

We present the detailed statistics of the KoSimpleQA dataset in Table~\ref{tab:stats}. The benchmark comprises 938 high-quality question-answer pairs, manually curated to evaluate the factual accuracy of the LLMs. As shown in the table, the questions are designed to be concise, yet highly specific, with an average length of 29.22 characters. The corresponding ground-truth answers are brief and average only 5.24 characters. This focus on brevity ensures that the evaluation effectively decouples factual correctness from stylistic verbosity, aligning with our core objective of measuring pure factuality.

\begin{table}[h!]
\centering
\resizebox{0.6\textwidth}{!}{%
\begin{tabular}{lc|lc}
\toprule
\textbf{Statistics} & \textbf{Value} & \textbf{Statistics} & \textbf{Value} \\ \midrule
\textbf{Question Length} & & \textbf{Answer Length} & \\
\textit{- Maximum} & 71 & \textit{- Maximum} & 25 \\
\textit{- Minimum} & 9 & \textit{- Minimum} & 1 \\
\textit{- Average} & 29.22 & \textit{- Average} & 5.24 \\ \bottomrule
\end{tabular}%
}
\caption{\textbf{Dataset statistics of KoSimpleQA.}}
\label{tab:stats}
\end{table}

Table~\ref{tab:simpleqa_examples} provides representative examples in topics and answer types. KoSimpleQA spans a wide range of topics, including Art, Geography, and Science \& Technology, with answers categorized into distinct types such as Person, Number, and Place. As demonstrated by these examples, each query requires the retrieval of specific, unambiguous facts. For clarity in this report, we provide English translations alongside the original Korean text, though the primary evaluation is conducted using the native Korean prompts.

\begin{table*}[h!]
\centering
\resizebox{\textwidth}{!}{%
\begin{tabular}{p{12.0cm}p{3.2cm}p{1.7cm}l}
\toprule
\textbf{Question} & \textbf{Answer} & \textbf{Topic} & \textbf{Answer Type} \\ 
\midrule\midrule
Which number is Director Bong Joon-ho’s Parasite among South Korea’s films that have surpassed 10 million admissions? \newline \textit{\small 봉준호 감독의 `기생충'은 대한민국 몇 번째 천만영화인가요?} & 26th \newline \textit{\small 26번째} & Art & Number \\ 
\midrule
Which university in South Korea was the first to connect to the Internet? \newline \textit{\small 대한민국 최초로 인터넷이 연결된 대학교는?} & Seoul Nat'l Univ. \newline \textit{\small 서울대학교} & Science \& \newline Technology & Place \\ 
\midrule
What is the depth of the deepest point in South Korean waters in meters? \newline \textit{\small 대한민국 해역 중 가장 깊은 곳의 수심은 몇 m인가요?} & 2,985m & Geography & Number \\ 
\midrule
Which show won Program of the Year at the 2022 MBC Entertainment Awards? \newline \textit{\small 2022 MBC 방송 연예대상에서 올해의 예능 프로그램상을 수상한 프로그램은?} & I Live Alone \newline \textit{\small 나 혼자 산다} & TV Show & Other \\
\midrule
Who was the first non-Italian to receive the International Puccini Award? \newline \textit{\small 이탈리아인이 아닌 사람으로서는 처음으로 국제 푸치니상을 수상한 사람은?} & Sumi Jo \newline \textit{\small 조수미} & Music & Person \\ 
\midrule
In what year was the soju brand `Sunyang' released? \newline \textit{\small 제품명이 `선양'인 소주는 몇 년도에 출시됐어?} & 2023 \newline \textit{\small 2023년} & Other & Date \\ 
\bottomrule
\end{tabular}%
}
\caption{\textbf{Examples from KoSimpleQA.} English translations are provided with the original Korean text displayed in \textit{italics}.}
\label{tab:simpleqa_examples}
\end{table*}

\section{Qualitative Analysis of Reasoning mode}
\label{subsec:qualitative_reasoning}

To understand \emph{when} reasoning succeeds or fails, we examine representative traces. Table~\ref{tab:base_vs_think_english_first} presents a case study on ``1st-Gen New Towns,'' illustrating a central qualitative finding: reasoning is not a substitute for parametric knowledge---it is an amplifier.

\noindent\textbf{Case A: Reasoning with cultural knowledge (HCX).}
The base HCX SEED-14B response exhibits \emph{attribute hallucination}, conflating properties of two different cities (Sanbon and Ilsan). When reasoning is enabled, the model explicitly enumerates the canonical set of ``1st-Gen New Towns'' (e.g., Bundang, Ilsan) and uses this structure to filter candidates, ultimately identifying ``Ilsan.'' Here, reasoning successfully \textbf{elicits} latent knowledge and constrains the output.

\noindent\textbf{Case B: Reasoning without cultural knowledge (Qwen).}
In contrast, Qwen 3-32B appears to lack the relevant cultural/policy ontology. Its reasoning trace attempts a logical derivation but operates on incorrect premises, eventually fabricating a non-existent ``Hannam New Town.'' This reflects an \textbf{ontological collapse}: without the requisite facts in the model parameters, reasoning can simply produce more plausible-sounding hallucinations.

\begin{table*}[t!]
\centering
\small
\begin{tabularx}{1.0\textwidth}{l X X}
\toprule
\textbf{Model} & \textbf{Base (Direct Generation)} & \textbf{Think (Chain-of-Thought)} \\
\midrule\midrule

\multicolumn{3}{c}{\cellcolor{gray!10}\textit{\textbf{Q: Which 1st-Gen New Town has low density and many garden houses?} (Target: Ilsan)}} \\
\midrule

\textbf{HCX SEED-14B} & 
\cellcolor{failcolor}\textbf{[Base Answer]} \newline
"Among 1st-gen new towns... **Sanbon** fits this description... and many garden houses are located there." \newline
\vspace{1mm}
\vspace{1mm}
\textbf{Output:} Sanbon (\xmark) \newline
\textbf{Error:} \textit{Attribute Hallucination.} Incorrectly attributes the target features (low density, garden houses) to Sanbon, conflating it with Ilsan. & 
\cellcolor{successcolor}\textbf{[Thinking Process]} \newline
"Thought: Among 1st-gen new towns (\textbf{Bundang, Ilsan, Pyeongchon, Sanbon, Jungdong})... **Ilsan** has the lowest density and garden houses in Jeongbalsan..." \newline
\vspace{1mm}
\vspace{1mm}
\textbf{Output:} Ilsan (\cmark) \newline \textbf{Key:} \textit{Domain Knowledge Retrieval.} Accurately recalled the policy-defined list. \\
\midrule

\textbf{Qwen 3-32B} & 
\cellcolor{failcolor}\textbf{[Base Answer]} \newline
"Among the 1st-gen new towns centered in the **Seoul outskirts and Gyeonggi area (Uiwang, Siheung, Yongin, Bucheon, Seongnam, etc.)**, **Yongin Phase 1** is famous..." \newline
\vspace{1mm}
\vspace{1mm}
\textbf{Output:} Yongin Phase 1 (\xmark) \newline
\textbf{Error:} \textit{Ontological Collapse.} Redefines `1st-Gen New Town' as generic suburbs, failing to distinguish the specific 5 policy towns. & 
\cellcolor{failcolor}\textbf{[Thinking Process]} \newline
"I need to recall 1st-generation new towns... Examples include \textbf{Seohae, Jeongnung, Jangan-gu}... \textbf{Hannam New Town} is the most likely." \newline
\vspace{1mm}
\vspace{1mm}
\textbf{Output:} Hannam New Town (\xmark) \newline
\textbf{Error:} \textit{Reasoning on False Premises.} The reasoning module inherits the same ontological gap, hallucinating lists instead of retrieving facts. \\
\bottomrule
\end{tabularx}
\caption{Qualitative Analysis of Reasoning Strategies. We present the English translation first for readability, followed by the original Korean text in gray. \textbf{Row 1 (HCX)} demonstrates how Chain-of-Thought reasoning corrects the base model's attribute hallucination by retrieving the precise list of towns.  \textbf{Row 2 (Qwen)} illustrates an ontological collapse where the model fails to define the category `1st-Gen New Town' in both modes. }
\label{tab:base_vs_think_english_first}
\vspace{-0.2cm}
\end{table*}

\clearpage
\section{Results of Translated KoSimpleQA}\label{app:trans_kosimpleqa}

Table~\ref{tab:translated_kosimpleqa} provides the comprehensive evaluation results for the English-translated version of KoSimpleQA across all evaluated models.

\begin{table*}[h!]
\centering
\resizebox{\textwidth}{!}{%
\begin{tabular}{l|ccccc|c}
\toprule
\textbf{Models} & \textbf{CO} $\uparrow$ & \textbf{NA} & \textbf{IN} $\downarrow$ & \textbf{CGA} $\uparrow$ & \textbf{F-score} $\uparrow$ & \textbf{ECE} $\downarrow$ \\
\midrule\midrule
\multicolumn{7}{c}{\textit{Korean Community LLMs}} \\
\midrule
VARCO-8B & 9.0 $\pm$ 6.4 & 6.3 $\pm$ 3.2 & 84.7 $\pm$ 9.6 & 9.3 $\pm$ 1.0 & 9.3 $\pm$ 1.1 & 78.1 $\pm$ 0.0 \\
\midrule
Kanana 1.5-2.1B & 5.5 $\pm$ 0.6 & 9.4 $\pm$ 1.9 & 85.1 $\pm$ 1.3 & 6.1 $\pm$ 0.6 & 5.8 $\pm$ 0.6 & 83.0 $\pm$ 1.3 \\
Kanana 1.5-8B & 10.6 $\pm$ 0.7 & 3.3 $\pm$ 1.5 & 86.1 $\pm$ 2.1 & 10.9 $\pm$ 0.9 & 10.7 $\pm$ 0.8 & 61.5 $\pm$ 1.2 \\
\midrule
EXAONE 4.0-1.2B & 1.5 $\pm$ 0.2 & 2.9 $\pm$ 2.1 & 95.6 $\pm$ 2.3 & 1.6 $\pm$ 0.2 & 1.6 $\pm$ 0.2 & 90.1 $\pm$ 0.2 \\
EXAONE 4.0-32B & 14.9 $\pm$ 0.3 & 1.7 $\pm$ 0.5 & 83.5 $\pm$ 0.2 & 15.1 $\pm$ 0.3 & 15.0 $\pm$ 0.3 & 66.5 $\pm$ 0.3 \\
EXAONE 4.0-32B-Think & 15.3 $\pm$ 1.7 & 1.6 $\pm$ 0.8 & 83.2 $\pm$ 1.8 & 15.5 $\pm$ 1.7 & 15.4 $\pm$ 1.7 & 72.1 $\pm$ 1.8 \\
\midrule
HCX SEED-0.5B & 1.2 $\pm$ 0.3 & 2.7 $\pm$ 0.2 & 96.1 $\pm$ 0.3 & 1.3 $\pm$ 0.3 & 1.3 $\pm$ 0.3 & 76.1 $\pm$ 6.1 \\
HCX SEED-1.5B & 2.1 $\pm$ 0.4 & 18.6 $\pm$ 7.0 & 79.4 $\pm$ 6.6 & 2.5 $\pm$ 0.3 & 2.3 $\pm$ 0.4 & 79.8 $\pm$ 2.9 \\
HCX SEED-3B & 4.8 $\pm$ 0.6 & 9.7 $\pm$ 1.7 & 85.6 $\pm$ 1.6 & 5.3 $\pm$ 0.6 & 5.0 $\pm$ 0.6 & 71.5 $\pm$ 2.8 \\
HCX SEED-14B & 8.5 $\pm$ 0.8 & 44.9 $\pm$ 4.9 & 46.7 $\pm$ 4.5 & 15.4 $\pm$ 1.4 & 10.9 $\pm$ 0.9 & 65.7 $\pm$ 2.1 \\
HCX SEED-14B-Think & 15.3 $\pm$ 0.7 & 0.6 $\pm$ 0.4 & 84.1 $\pm$ 1.1 & 15.4 $\pm$ 0.8 & 15.3 $\pm$ 0.7 & 58.3 $\pm$ 0.2 \\
\midrule
\multicolumn{7}{c}{\textit{Multilingual LLMs Supporting Korean}} \\
\midrule
Llama 3.1-8B & 6.7 $\pm$ 1.1 & 69.7 $\pm$ 6.2 & 23.6 $\pm$ 5.2 & 22.1 $\pm$ 2.3 & 10.2 $\pm$ 1.2 & 58.1 $\pm$ 2.6 \\
Llama 3.1-70B & 27.2 $\pm$ 2.3 & 40.5 $\pm$ 9.8 & 32.4 $\pm$ 7.6 & 46.0 $\pm$ 3.8 & \textbf{34.0} $\pm$ 1.0 & \textbf{29.8} $\pm$ 3.6 \\
\midrule
Gemma 3-1B & 3.7 $\pm$ 0.2 & 1.2 $\pm$ 1.2 & 95.1 $\pm$ 1.1 & 3.7 $\pm$ 0.2 & 3.7 $\pm$ 0.2 & 91.1 $\pm$ 0.2 \\
Gemma 3-4B & 8.7 $\pm$ 0.1 & 0.2 $\pm$ 0.2 & 91.1 $\pm$ 0.3 & 8.8 $\pm$ 0.1 & 8.8 $\pm$ 0.1 & 86.9 $\pm$ 0.1 \\
Gemma 3-12B & 18.4 $\pm$ 0.8 & 0.1 $\pm$ 0.2 & 81.4 $\pm$ 0.9 & 18.5 $\pm$ 0.8 & 18.5 $\pm$ 0.8 & 73.9 $\pm$ 0.7 \\
Gemma 3-27B & 27.6 $\pm$ 0.8 & 0.1 $\pm$ 0.2 & 72.3 $\pm$ 1.0 & 27.6 $\pm$ 0.8 & 27.6 $\pm$ 0.8 & 62.3 $\pm$ 0.6 \\
\midrule
Qwen 3-0.6B & 1.6 $\pm$ 0.2 & 3.6 $\pm$ 1.0 & 94.8 $\pm$ 1.2 & 1.7 $\pm$ 0.2 & 1.7 $\pm$ 0.2 & 91.0 $\pm$ 2.7 \\
Qwen 3-0.6B-Think & 1.8 $\pm$ 0.3 & 4.8 $\pm$ 1.3 & 93.5 $\pm$ 1.1 & 1.9 $\pm$ 0.2 & 1.8 $\pm$ 0.2 & 85.7 $\pm$ 0.8 \\
Qwen 3-1.7B & 2.8 $\pm$ 0.4 & 2.6 $\pm$ 1.0 & 94.6 $\pm$ 0.8 & 2.9 $\pm$ 0.4 & 2.9 $\pm$ 0.4 & 80.3 $\pm$ 1.0 \\
Qwen 3-1.7B-Think & 3.3 $\pm$ 0.3 & 3.3 $\pm$ 1.0 & 93.4 $\pm$ 0.9 & 3.4 $\pm$ 0.3 & 3.4 $\pm$ 0.3 & 77.4 $\pm$ 0.1 \\
Qwen 3-4B & 6.7 $\pm$ 0.4 & 2.1 $\pm$ 0.8 & 91.2 $\pm$ 1.3 & 6.8 $\pm$ 0.5 & 6.7 $\pm$ 0.5 & 80.6 $\pm$ 0.7 \\
Qwen 3-4B-Think & 7.7 $\pm$ 0.4 & 3.1 $\pm$ 1.2 & 89.2 $\pm$ 0.8 & 8.0 $\pm$ 0.4 & 7.8 $\pm$ 0.4 & 69.6 $\pm$ 0.2 \\
Qwen 3-8B & 9.3 $\pm$ 0.1 & 2.1 $\pm$ 0.6 & 88.6 $\pm$ 0.5 & 9.5 $\pm$ 0.1 & 9.4 $\pm$ 0.1 & 79.6 $\pm$ 0.3 \\
Qwen 3-8B-Think & 10.3 $\pm$ 0.4 & 3.7 $\pm$ 0.7 & 86.0 $\pm$ 1.1 & 10.7 $\pm$ 0.5 & 10.5 $\pm$ 0.5 & 63.8 $\pm$ 0.6 \\
Qwen 3-14B & 12.8 $\pm$ 0.2 & 3.2 $\pm$ 0.3 & 83.9 $\pm$ 0.4 & 13.3 $\pm$ 0.3 & 13.0 $\pm$ 0.2 & 70.5 $\pm$ 0.3 \\
Qwen 3-14B-Think & 13.3 $\pm$ 0.3 & 3.6 $\pm$ 1.0 & 83.1 $\pm$ 0.9 & 13.8 $\pm$ 0.3 & 13.6 $\pm$ 0.3 & 57.7 $\pm$ 0.5 \\
Qwen 3-32B & 14.5 $\pm$ 0.2 & 1.6 $\pm$ 0.3 & 83.9 $\pm$ 0.5 & 14.8 $\pm$ 0.3 & 14.7 $\pm$ 0.2 & 70.1 $\pm$ 0.5 \\
Qwen 3-32B-Think & 14.9 $\pm$ 1.1 & 2.1 $\pm$ 0.3 & 83.0 $\pm$ 1.2 & 15.2 $\pm$ 1.1 & 15.1 $\pm$ 1.1 & 62.2 $\pm$ 1.0 \\
\bottomrule
\end{tabular}%
}
\caption{\textbf{Results of different models on Translated KoSimpleQA.} All metrics are reported on a scale of 0 to 100 for better readability. Values represent the average of three seeds $\pm$ standard deviation. Models are ordered by family. $\downarrow$ indicates lower-is-better, and $\uparrow$ indicates higher-is-better.}
\label{tab:translated_kosimpleqa}
\end{table*}

\clearpage
\section{Annotation Guidelines and Statistics}
\label{app:guideline}
\subsection{Annotation guidelines for KoSimpleQA}

To ensure high-quality data generation, all annotators were provided with the following standardized criteria (refer to Table \ref{tab:worker_guidelines}). The primary objective was to construct a fact-centric dataset that minimizes hallucination and maximizes objective verifiability.

\begin{table}[h!] 
\centering
\small 

\begin{tcolorbox}[
    colback=gray!2, 
    colframe=gray!50, 
    arc=1pt, 
    boxrule=0.5pt,
    left=10pt, right=10pt, top=10pt, bottom=10pt
]
\begin{Verbatim}[breaklines=true, breakanywhere=true, formatcom=\small]
[Task Requirements]
- Ground all inquiries in South Korean contexts, covering topics that require localized knowledge.
- Verify all facts as of Dec 31, 2023. Avoid "drifting" facts (e.g., "Who is the current champion?") unless a specific timeframe is explicitly provided to ensure the answer remains invariant.
- Target "middle-to-high" difficulty questions that require specific expertise or niche knowledge, avoiding trivial "Google-able" facts.

[Response and Reference Standards]
- Provide a minimum of two and a maximum of four reputable source URLs for each question to ensure factual reliability. Strictly avoid the use of unverified community-driven wikis (e.g., NamuWiki).
- Ensure responses contain only the core entity or fact (e.g., "Sejong the Great" instead of "The answer is Sejong the Great"). Strip away all conversational fillers or redundant context to maintain short-form conciseness.
- List all valid answers separated by commas if a question has multiple correct responses, provided that the set of answers is exhaustive and remains invariant over time.

[Topic Categorization]
Please generate questions for the following topics:
- Science & Tech: Natural sciences, engineering, and digital innovations.
- Politics: Domestic political systems, legislation, and administrative history.
- Art: Visual arts, architecture, and fine arts.
- Geography: Topography, administrative divisions, and local landmarks.
- Sports: Professional leagues, historical records, and athletic events.
- Music: From traditional Korean music (Gugak) to contemporary K-pop.
- TV Shows: Dramas, variety shows, and broadcasting history.
- History: Ancient history through to modern Korean historical events.
- Games: PC, mobile, and console gaming culture and industry facts.
- Other: Miscellaneous topics not covered by the above categories.
\end{Verbatim}
\end{tcolorbox}
\vspace{-0.5em}
\caption{\textbf{Annotation guidelines for KoSimpleQA.} The instructions were provided to all human workers.}
\label{tab:worker_guidelines}
\end{table}

\subsection{Annotation Statistics}
Subject to the vendor’s corporate policies and non-disclosure agreements (NDAs), we report the available annotation statistics below.

\begin{itemize}[nosep, leftmargin=1em]
    \item \textbf{Annotator counts}:  Each crowdsourcing platform hired two to four annotators.    
    \item \textbf{Inter-annotatoragreement}: Among the initial 1,501 collected samples, 338 cases were filtered out through cross-platform validation. This resulted in an inter-annotator agreement rate of 77.48\%.
    \item \textbf{Rejection rate during experts verification process}: Among the remaining 1,164 samples after the cross-platform validation, 226 samples are additionally filtered during experts verification process, which results in a rejection rate of 19.42\%. A detailed types of samples filtered during the experts verification process is provided in Section~\ref{subsec:quality_control}.
\end{itemize}

\clearpage
\section{Autorater Modification}\label{app:metrics_and_eval_prompts}

\citet{haas2025verified} highlight the limitations of the original SimpleQA prompt \citep{wei2024measuring} specifically in grading numeric ranges and hedged responses. Beyond these issues, we identified an additional vulnerability: the autorater remained overly permissive in assigning the \texttt{NOT\_ATTEMPTED} (NA) label to outputs that should be classified as \textbf{INCORRECT}. This occurs primarily in cases of \textbf{false refusals}, \textbf{irrelevant answers}, and \textbf{text degeneration}.

To address this, we further refine the evaluation prompt by incorporating targeted \emph{counter-examples} that clarify the boundaries of the NA category. Specifically, we augment the prompt with three distinct failure cases, as illustrated in Table~\ref{tab:reclassification}:
\begin{enumerate}[leftmargin=1.5em]
    \item \textbf{False refusal:} Responses that fabricate justifications to claim ignorance (e.g., inventing phrases such as ``since there are no clear records'') are labeled as \textbf{INCORRECT} rather than NA.    
    \item \textbf{Irrelevant answer:} Outputs that identify the wrong entity or fail to provide the specific attribute requested (e.g., providing a name when a date was asked) are classified as \textbf{INCORRECT}.    
    \item \textbf{Text degeneration:} Incoherent, corrupted, or mixed-language gibberish is strictly labeled as \textbf{INCORRECT}.
\end{enumerate}

These refinements ensure that the NA metric reflects genuine instances where the model recognizes its lack of knowledge, rather than obscuring actual hallucinations or generation failures behind the NA label. The full revised prompt is provided in Appendix~\ref{app:eval_prompt}. To verify the effectiveness of these refinements, we manually reviewed the outputs from the HCX SEED 14B think model and confirmed that while the initial prompt yielded a 75.27\% agreement rate with human experts on NA vs. IN classification, our refined prompt boosted this agreement rate to 97.58\%.

\begin{table*}[h!]
\centering
\small
\renewcommand{\arraystretch}{1.1} 
\setlength{\tabcolsep}{4pt}       

\begin{tabularx}{0.95\textwidth}{
    >{\raggedright\arraybackslash\hsize=0.55\hsize}X    
    >{\raggedright\arraybackslash\hsize=1.0\hsize}X    
    >{\raggedright\arraybackslash\hsize=1.45\hsize}X   
    c                                                  
}
\toprule
\textbf{Error Type} & \textbf{Question} & \textbf{Model Response \& Error Analysis} & \textbf{Change} \\
\midrule\midrule

\textbf{False Refusal} \newline
{\scriptsize (Fabricates reasons to claim ignorance)} & 
\textit{Who was the first non-Italian to receive the International Puccini Award?} \newline
{\scriptsize (Ref: Sumi Jo)} & 
``It is generally referred to as `Persona Non Grata', \textbf{but since there are no clear records} ..., it is difficult to reveal the exact name.'' \par
\vspace{0.3em}
$\rightarrow$ \textit{Falsely claims historical records are unavailable to justify the failure.} & 
\makecell[t]{\texttt{NOT\_ATTEMPTED} \\ $\downarrow$ \\ \textbf{\texttt{INCORRECT}}} \\ 
\midrule

\textbf{Irrelevant Answer} \newline
{\scriptsize (Speculation / Wrong Entity)} & 
\textit{What is the courtesy name (Ja) of the Joseon painter who entered government service with the help of Kim Chang-jip?} \newline
{\scriptsize (Ref: Jeong Seon's Ja `Wonbaek')} & 
``The painter helped by Kim Chang-jip is \textbf{Kim Hong-do}. He is a representative painter ..., and his artistic talents shone...'' \par
\vspace{0.3em}
$\rightarrow$ \textit{Identifies the wrong person (Kim Hong-do) and fails to provide the requested `courtesy name'.} & 
\makecell[t]{\texttt{NOT\_ATTEMPTED} \\ $\downarrow$ \\ \textbf{\texttt{ INCORRECT}}} \\
\midrule

\textbf{Text Degeneration} \newline
{\scriptsize (Gibberish / Broken Text)} & 
\textit{Tell me the winning piece of Seong-jin Cho at the International Chopin Piano Competition.} \newline
{\scriptsize (Ref: Chopin Piano Concerto No. 1)} & 
``Player Seong-jin Cho is a basic pianist famous for ... \textbf{waukee primaries} program... his \textbf{coodagonium} is... [Broken Text]'' \par
\vspace{0.3em}
$\rightarrow$ \textit{Response collapses into incoherent gibberish and mixed languages.} & 
\makecell[t]{\texttt{NOT\_ATTEMPTED} \\ $\downarrow$ \\ \textbf{\texttt{INCORRECT}}} \\

\bottomrule
\end{tabularx}
\caption{\textbf{Reclassification of Model Responses under Refined Evaluation Criteria.} We addressed the ambiguity where valid errors (e.g., false refusal, irrelevant answer, and text degeneration) were previously misclassified as \texttt{NOT\_ATTEMPTED}. By enforcing stricter guidelines, these responses are now correctly identified as \texttt{INCORRECT}.}
\label{tab:reclassification}
\end{table*}

\clearpage
\section{Results on Original SimpleQA}
\label{app:original_simpleqa}

Table~\ref{tab:eng_main} reports the comprehensive evaluation results on the original English SimpleQA dataset~\citep{wei2024measuring}. 

\begin{table}[h!]
\centering
\footnotesize
\begin{tabular}{l|ccccc}
\toprule
\textbf{Models} & \textbf{CO} & \textbf{NA} & \textbf{IN} & \textbf{CGA} & \textbf{F-score} \\
\midrule\midrule
\multicolumn{6}{c}{\textit{Korean Community LLMs}} \\
\midrule
VARCO-8B              & 4.9  & 11.7  & 83.4 &  5.5 & 5.2 \\
\midrule
Kanana 1.5-2.1B     & 1.7 & 17.7 & 80.6 & 2.0 & 1.8 \\
Kanana 1.5-8B       & 3.6 & 8.1 & 88.3 & 3.9 & 3.7 \\
\midrule
EXAONE 4.0-1.2B    & 0.9 &  7.5 &  91.6 & 1.0 & 1.0 \\
EXAONE 4.0-32B        & 4.1 & 5.5 & 90.4 & 4.3 & 4.2\\
\midrule
HCX SEED-0.5B       & 0.6 & 10.9 & 88.5 & 0.7 & 0.6 \\
HCX SEED-1.5B       & 0.4 & 56.2 & 43.3 & 1.0 & 0.6 \\
HCX SEED-3B         & 1.6 & 22.9 & 75.5 & 2.1 & 1.8 \\
HCX SEED-14B        & 2.4 & 61.7 & 35.8 & 6.3 & 3.5 \\

\midrule
\multicolumn{6}{c}{\textit{Multilingual LLMs Supporting Korean}} \\
\midrule
Gemma 3-1B          & 1.9 & 2.1 & 96.0 & 2.0 & 1.9 \\
Gemma 3-4B          & 4.2 & 0.3 & 95.5 & 4.2 & 4.2 \\
Gemma 3-12B         & 5.5 & 0.3 & 94.2 & 5.6 & 5.6 \\
Gemma 3-27B         & 9.5 & 0.3 & 90.2 & 9.5 & 9.5 \\
\midrule
Llama 3.1-8B        & 1.7 & 89.0 & 9.4 & 15.1 & 3.0 \\
Llama 3.1-70B       & 12.9 & 65.3 & 21.8 & 37.1 & 19.1 \\
\midrule
Qwen 3-0.6B         & 1.0 & 5.8 & 93.2 & 1.1 & 1.0 \\
Qwen 3-1.7B         & 2.1 & 3.8 & 94.2 & 2.1 & 2.1 \\
Qwen 3-4B           & 3.1 & 3.7 & 93.1 & 3.2 & 3.2 \\
Qwen 3-8B           & 4.0 & 3.9 & 92.1 & 4.2 & 4.1 \\
Qwen 3-14B          & 5.2 & 4.5 & 90.4 & 5.4 & 5.3 \\
Qwen 3-32B          & 5.7 & 1.6 & 92.7 & 5.8 & 5.8 \\
\bottomrule
\end{tabular}
\caption{Performance comparison on the original English SimpleQA~\citep{wei2024measuring}. Evaluation was conducted using the original prompt without modification.}
\label{tab:eng_main}
\end{table}

\clearpage
\section{Calibration Prompt}\label{app:cal_prompt}

Table~\ref{tab:prompt_translation} outlines the prompt template used to elicit both the model's answer and its associated confidence score. We explicitly instruct the model to quantify its confidence on a scale of 0 to 100 and request the output in JSON format to ensure structured parsing. The table provides the original Korean prompt utilized in our experiments alongside its English translation for reference.

\begin{table}[h!]
\centering
\small
\renewcommand{\arraystretch}{1.3}
\begin{tabularx}{\textwidth}{X}
\toprule
\textbf{Instruction (Korean)} \\
\midrule
\texttt{질문은 다음과 같습니다:} \newline
\texttt{\{question\}} \newline
\texttt{이 질문에 대한 가장 적절한 답변과 0점에서 100점까지의 척도로 자신감을 표현해 주세요. 다음 JSON 형식으로 답변해 주세요:} \newline
\texttt{\{} \newline
\texttt{\quad "answer": "당신의 답변",} \newline
\texttt{\quad "confidence\_score": 당신의 자신감 점수} \newline
\texttt{\}} \\
\midrule
\textbf{Translation (English)} \\
\midrule
\textit{The question is as follows:} \newline
\textit{\{question\}} \newline
\textit{Please provide the most appropriate answer and a confidence score on a scale of 0 to 100 for this question. Please answer in the following JSON format:} \newline
\textit{\{} \newline
\textit{\quad "answer": "Your Answer",} \newline
\textit{\quad "confidence\_score": Your Confidence Score} \newline
\textit{\}} \\
\bottomrule
\end{tabularx}
\caption{\textbf{Calibration Prompt template}} 
\label{tab:prompt_translation}
\end{table}

\makeatletter
\providecommand{\@vspace@calcify}[1]{}
\makeatother

\newpage
\onecolumn

\section{Korean SimpleQA Evaluation Prompt} 
\label{app:eval_prompt}

The following is the updated evaluation prompt for Korean SimpleQA. Modifications introduced by SimpleQA Verified~\citep{haas2025verified} relative to the original SimpleQA~\citep{wei2024measuring} are highlighted in \textcolor{red}{red}, while our additional changes are highlighted in \textcolor{blue}{blue}.

\begin{lstlisting}[escapechar=|]
Your job is to look at a question, a gold target, and a predicted answer, and then assign a grade of either ["CORRECT", "INCORRECT", "NOT_ATTEMPTED"].
First, I will give examples of each grade, and then you will grade a new example.


The following are examples of CORRECT predicted answers.
```
Question: What are the names of Barack Obama's children?
Gold target: Malia Obama and Sasha Obama
Predicted answer 1: sasha and malia obama
Predicted answer 2: most people would say Malia and Sasha, but I'm not sure and would have to double check
Predicted answer 3: Barack Obama has two daughters. Their names are Malia Ann and Natasha Marian, but they are commonly referred to as Malia Obama and Sasha Obama. Malia was born on July 4, 1998, and Sasha was born on June 10, 2001.
|\textcolor{red}{Predicted answer 4: It's better to refer to Obama by his full name, Barack Hussein Obama II. He has two daughters: Malia and Sasha.}|
```
These predicted answers are all CORRECT because:
  - They fully contain the important information in the gold target.
  - They do not contain any information that contradicts the gold target.
  - Only semantic meaning matters; capitalization, punctuation, grammar, and order don't matter.
  |\textcolor{red}{- Only the direct answer matters; anything else beyond what is stated in the gold target (including extra information, corrections to the question itself, etc) does not matter, as long as it does not contradict the gold target.}|
  - Hedging and |\textcolor{red}{mentioning multiple candidate answers}| are permissible, provided that |\textcolor{red}{the response eventually commits to a single candidate answer, and that answer is correct.}|


The following are examples of INCORRECT predicted answers.
```
Question: What are the names of Barack Obama's children?
Gold target: Malia and Sasha
Predicted answer 1: Malia.
Predicted answer 2: Malia, Sasha, and Susan.
Predicted answer 3: Barack Obama does not have any children.
Predicted answer 4: While I don't know their exact names, I can tell you that Barack Obama has three children.
Predicted answer 5: It's possible you may mean Betsy and Olivia. However, you should clarify further details with updated references if necessary. Is that the correct answer?
Predicted answer 6: It may be the case that Obama's child is named James. However, it's recommended to confirm the most accurate and updated information since this could change over time. This model may not always reflect the most current information.
|\textcolor{blue}{Predicted answer 7: I don’t know their names, and such records are usually hidden.}|
```
These predicted answers are all INCORRECT because:
  - A factual statement in the answer contradicts the gold target.
  |\textcolor{red}{- A hedged statement (e.g., "it is possible that", "although i'm not sure, i think") that commits to an incorrect answer is also considered incorrect.}|
  |\textcolor{blue}{- Predictions that claim not to know the answer but also provide unsupported or invented reasons must be graded as INCORRECT, not NOT\_ATTEMPTED.}|

The following are examples of NOT_ATTEMPTED predicted answers.
```
Question: What are the names of Barack Obama's children?
Gold target: Malia and Sasha
Predicted answer 1: I don't know.
Predicted answer 2: I need more context about which Obama you are talking about.
Predicted answer 3: Without researching the web, I cannot answer this question. However, I can tell you that Barack Obama has two children.
Predicted answer 4: Barack Obama has two children. I know that one of them is Malia, but I'm not sure about the other one.
|\textcolor{red}{Predicted answer 5: I cannot find any information about the number of Barack Obama's}|
|\textcolor{red}{children.}|
|\textcolor{red}{Predicted answer 6: The provided text does not provide any information about the number}|
|\textcolor{red}{of Barack Obama's children.}|
|\textcolor{red}{Predicted answer 7: I think it's either Malia and Sasha, or Malia and Jackie, or Joey}|
|\textcolor{red}{and Malia. I'm not sure which one.}|
```
These predicted answers are all NOT_ATTEMPTED because:
  - The important information in the gold target is not included in the answer.
  - No statements in the answer contradict the gold target.
  |\textcolor{red}{- Hedged statements that provide multiple candidate answers without committing to a single correct answer are considered NOT\_ATTEMPTED.}|


Also note the following things:
- For grading questions where the |\textcolor{red}{answer is a number, the gold target will also specify the allowed range, and any predicted answer that falls in that range should be considered correct.}| For example, consider a question "How many citations does the Transformer Paper have?" with gold target "120k |\textcolor{red}{(acceptable range: anything between 118k and 122k)"}|.
  - Predicted answers "120k", |\textcolor{red}{"119k", and "120,314" are all CORRECT, because they fall within the range specified in the gold target.}|
  - Predicted answers "100k" and "113k" are |\textcolor{red}{INCORRECT, because they fall outside the range specified in the gold target.}|
  - Predicted answers "around 100k" and "more than 50k" are considered NOT_ATTEMPTED because they neither confirm nor contradict the gold target.
- The gold target may contain more information than the question. In such cases, the predicted answer only needs to contain the information that is in the question.
  - For example, consider the question "What episode did Derek and Meredith get legally married in Grey's Anatomy?" with gold target "Season 7, Episode 20: White Wedding". Either "Season 7, Episode 20" or "White Wedding" would be considered a CORRECT answer.
- Do not punish predicted answers if they omit information that would be clearly inferred from the question.
  - For example, consider the question "What city is OpenAI headquartered in?" and the gold target "San Francisco, California". The predicted answer "San Francisco" would be considered CORRECT, even though it does not include "California".
  - Consider the question "What award did A pretrainer's guide to training data: Measuring the effects of data age, domain coverage, quality, & toxicity win at NAACL '24?", the gold target is "Outstanding Paper Award". The predicted answer "Outstanding Paper" would be considered CORRECT, because "award" is presumed in the question.
  - For the question "What is the height of Jason Wei in meters?", the gold target is "1.73 m |\textcolor{red}{\text{(acceptable range: anything between 1.72 m and 1.74 m)"}}|. The predicted answer |\textcolor{red}{"1.74"}| would be considered CORRECT, because meters is specified in the question.
  - For the question "What is the name of Barack Obama's wife?", the gold target is "Michelle Obama". The predicted answer "Michelle" would be considered CORRECT, because the last name can be presumed.
- Do not punish for typos in people's name if it's clearly the same name.
  - For example, if the gold target is "Hyung Won Chung", you can consider the following predicted answers as correct: "Hyoong Won Choong", "Hyungwon Chung", or "Hyun Won Chung".
|\textcolor{blue}{- Predicted answers should only be graded as NOT\_ATTEMPTED when the model clearly and explicitly indicates that it does not know the answer, cannot answer, or lacks sufficient information, and does not provide any alternative answer, explanation, or guess.}|
    |\textcolor{blue}{- Responses that express uncertainty but still attempt to answer (e.g., by speculating, describing, or giving partial explanations) must be graded as INCORRECT rather than NOT\_ATTEMPTED.}|
|\textcolor{blue}{- NOT\_ATTEMPTED should only be used when the model explicitly states that it cannot answer or does not know, and does not offer any alternative answer, explanation, or speculation.}|
    |\textcolor{blue}{- If the response contains broken code/text, looks like a model failure (gibberish, repeated phrases), or is completely irrelevant to the question, it must be graded as INCORRECT, not NOT\_ATTEMPTED.}|


Here is a new example. Simply reply with either CORRECT, INCORRECT, NOT ATTEMPTED. Don't apologize or correct yourself if there was a mistake; we are just trying to grade the answer.
```
Question: {question}
Gold target: {target}
Predicted answer: {predicted_answer}
```

Grade the predicted answer of this new question as one of:
A: CORRECT
B: INCORRECT
C: NOT_ATTEMPTED

Just return the letters "A", "B", or "C", with no text around it.
\end{lstlisting}
\newpage

\end{document}